\documentclass[11pt,reqno]{amsart} 
\usepackage[margin=1.1in]{geometry}
\usepackage{amsthm, amsfonts, amssymb, amsmath,enumitem, bbm, mathabx,booktabs,multirow}
\usepackage{mathtools}
\usepackage{mathrsfs} 
\usepackage{algorithm}
\usepackage{algorithmic}
\usepackage{pgf,tikz,pgfplots} 
\usetikzlibrary{arrows}
\usepackage{enumitem}
\usepackage{comment}
\usepackage{chngcntr}
\usepackage{mathabx}
\usepackage{bm}
\usepackage{etoolbox}
\usepackage{float}
\usepackage{subcaption}
\usepackage{pkgfile}

\usepackage{abraces} 
\usepackage{eufrak}
\usepackage{upgreek} 
\usepackage{booktabs} 
\usepackage{array} 
\usepackage{verbatim}

\usepackage{natbib}

\newtheorem{theorem}{Theorem}[section]

\newtheorem{proposition}{Proposition}[section]

\newtheorem{lemma}{Lemma}[section]

\theoremstyle{definition} 
\newtheorem{definition}{Definition}[section]
\newtheorem{remark}{Remark}[section]

\newcommand{\wb}{\bar}

\newcommand{\E}{\mathbb E}

\usepackage[colorlinks=true, 
            bookmarksnumbered=true,
			linkcolor=black, 
			citecolor=blue, 
			urlcolor=blue]{hyperref}

\allowdisplaybreaks

\begin{document}

\title[Private $\beta$ Models]{Privacy utility trade offs for parameter estimation in degree-heterogeneous higher-order networks}

\author{Bibhabasu Mandal}
\address{Indian Statistical Institute, Kolkata, India}
\email{bibhabasumandal04@gmail.com}
\thanks{Both authors contributed equally to this work.}

\author{Sagnik Nandy}
\address{Department of Statistics, The Ohio State University, Columbus, OH 43210, USA}
\email{nandy15@osu.edu}
\thanks{\protect\footnotemark}

\begin{abstract}
In sensitive applications involving relational datasets, protecting information about individual links from adversarial queries is of paramount importance. In many such settings, the available data are summarized solely through the degrees of the nodes in the network. We adopt the $\beta$-model, which is the prototypical statistical model adopted for this form of aggregated relational information, and study the problem of minimax-optimal parameter estimation under both local and central differential privacy constraints. We establish finite-sample minimax lower bounds that characterize the precise dependence of the estimation risk on the network size and the privacy parameters, and we propose simple estimators that achieve these bounds up to constants and logarithmic factors under both local and central differential privacy frameworks. Our results provide the first comprehensive finite-sample characterization of privacy–utility trade-offs for parameter estimation in $\beta$-models, addressing the classical graph case and extending the analysis to higher-order hypergraph models. We further demonstrate the effectiveness of our methods through experiments on synthetic data and a real-world communication network.

\end{abstract}

\maketitle

\section{Introduction}
Protecting the privacy of individual links while analyzing network data is a fundamental challenge in modern data analysis, particularly in the presence of adversarial queries that may exploit released information to infer sensitive user attributes. Naïve approaches such as anonymizing network nodes \citep{hay2007anonymizing} are now well understood to be insufficient: carefully designed re-identification attacks can often recover private information \citep{narayanan2009anonymizing}. The increasing ubiquity of relational data, arising from social interactions, financial transactions, communication patterns, and contact-tracing networks—has further heightened the need for principled mechanisms that allow the release of aggregate or structural network information while rigorously protecting the privacy of individuals.

These concerns are especially acute in networks involving human subjects, where the presence or absence of a single link may encode highly sensitive information, such as participation in a financial transaction or exposure to a disease. Differential privacy, introduced in the seminal work of \cite{dwork2014algorithmic}, has emerged as the benchmark framework for quantifying privacy guarantees in such settings. A substantial literature has since developed mechanisms for differentially private data release and inference in network and relational data, addressing challenges posed by dependence, correlation, and adversarial querying \citep{kasiviswanathan2013analyzing,nissim2007smooth,nguyen2016detecting,mulle2015privacy,nguyen2015differentially,chen2014correlated}.

In many sensitive applications—such as contact tracing, financial transactions, and social media–based marketing—the only information released by data providers is the total number of connections associated with each node in the network. For instance, epidemiological studies of sexually transmitted diseases often collect data only on the number of sexual partners reported by each individual \citep{helleringer2007sexual}. While such compression avoids explicit disclosure of individual relationships, degree information may still encode sensitive demographic, geographic, or community-level attributes, which can be exploited by adversarial analysts without the consent of the individuals under study. This motivates the development of robust mechanisms for releasing degree-based statistics under formal differential privacy guarantees.

Moreover, in many modern settings, relational data are inherently higher order rather than pairwise. Examples include folksonomies \citep{ghoshal2009random}, citation and coauthorship networks \citep{ji2016coauthorship,patania2017shape,karwa2016discussion}, social or professional group interactions \citep{social_groups_hypergraphs,Benson-2018-simplicial}, and biological networks \citep{grilli2017higher,michoel2012alignment,petri2014homological}. In such cases, the natural representation is a hypergraph, and the available data often consist solely of node-level participation counts in higher-order interactions. A central question then is how to release or analyze these degree-based summaries while ensuring differential privacy.

\subsection{Statistical model and problem formulation}

When relational data are available only in summarized form—where the number of interactions in which each individual participates is exposed—a natural modeling choice is the $\beta$-model introduced by \cite{park2004statistical}, which is a special case of the $p_1$-model of \cite{holland1981exponential}. In this model, the observed degree sequence is assumed to be generated by a random graph $\mathcal G=(\mathcal V,\mathcal E)$ with vertex set $\mathcal V=[n]:=\{1,\ldots,n\}$, where each edge $(i,j)$ is independently included in $\mathcal E$ with probability
\[
\mathbb P\left[(i,j)\in\mathcal E\right]
=\frac{e^{\beta_i+\beta_j}}{1+e^{\beta_i+\beta_j}},
\quad \text{for some } \beta\in\mathbb R^n .
\]
The degree sequence $d:=(d_1,\ldots,d_n)$, where $d_i=\sum_{j\neq i}\mathbbm 1\{(i,j)\in\mathcal E\}$, is a sufficient statistic for $\beta$, making the model appropriate for settings in which only degree information is available.

The graph $\beta$-model has been extended to hypergraphs in \cite{Stasi2014MF}. In the $r$-uniform hypergraph $\beta$-model, the data are assumed to be generated from a random hypergraph $\mathcal G=(\mathcal V,\mathcal E_r)$ with vertex set $\mathcal V=[n]$, where each hyperedge $(i_1,\ldots,i_r)$ (composed of a tuple of $r$ distinct entries from $[n]$) is independently included in $\mathcal E_r$ with probability
\begin{align}
\label{eq:r_uniform_beta_model}
\mathbb P\left[(i_1,\ldots,i_r)\in\mathcal E_r\right]
=\frac{e^{\beta_{i_1}+\cdots+\beta_{i_r}}}{1+e^{\beta_{i_1}+\cdots+\beta_{i_r}}},
\quad \beta\in\mathbb R^n .
\end{align}
In this case, the sufficient statistics are the $r$-degrees $(d_{r,1},\ldots,d_{r,n})$, where $d_{r,i}=\sum_{e\in\mathcal E_r}\mathbbm 1\{i\in e\}$.

In the $r$-uniform hypergraph $\beta$-model, a standard approach to parameter estimation is maximum likelihood. 
The graph $\beta$-model corresponds to the special case $r=2$. In both the graph and hypergraph settings, the statistical properties of the MLE $\hat\beta$ have been extensively studied under various asymptotic regimes \citep{10.1214/12-AOS1078,2010arXiv1005.1136C,Stasi2014MF,nandy_bhattacharya}.

Despite this progress, relatively few works have addressed privacy-preserving release of degree information or parameter estimation in the $\beta$-model. In \cite{karwa_et_al}, the authors proposed a differentially private estimation procedure based on solving the maximum likelihood optimization using noisy degree statistics. Their analysis, however, is restricted to the graph case and focuses on asymptotic regimes under a specific scaling of the privacy parameter, without providing finite-sample characterizations of the privacy-utility trade-off. More recently, \cite{10.1214/24-AOS2365} studied private method-of-moments estimators in the $\beta$-model and established finite-sample $\ell_\infty$ risk bounds, but their method relies on observing the actual connections. Furthermore, their bounds do not capture the precise dependence of the estimation error on the privacy parameter, nor establish minimax optimality.

To the best of our knowledge, there has been no principled study that characterizes the exact worst-case finite sample privacy-utility trade-off for parameter estimation in the $\beta$-model. In this paper, we seek to fill this gap by addressing the following questions:
\begin{enumerate}
\item What is the optimal differentially private mechanism for releasing degree information?
\item Given a trusted curator who observes the raw degrees and releases privatized aggregate statistics, what is the optimal way to construct such statistics for parameter estimation?
\end{enumerate}

\begin{figure}[t]
  \centering
  \begin{minipage}[t]{0.48\linewidth}
    \centering
    \includegraphics[width=\linewidth]{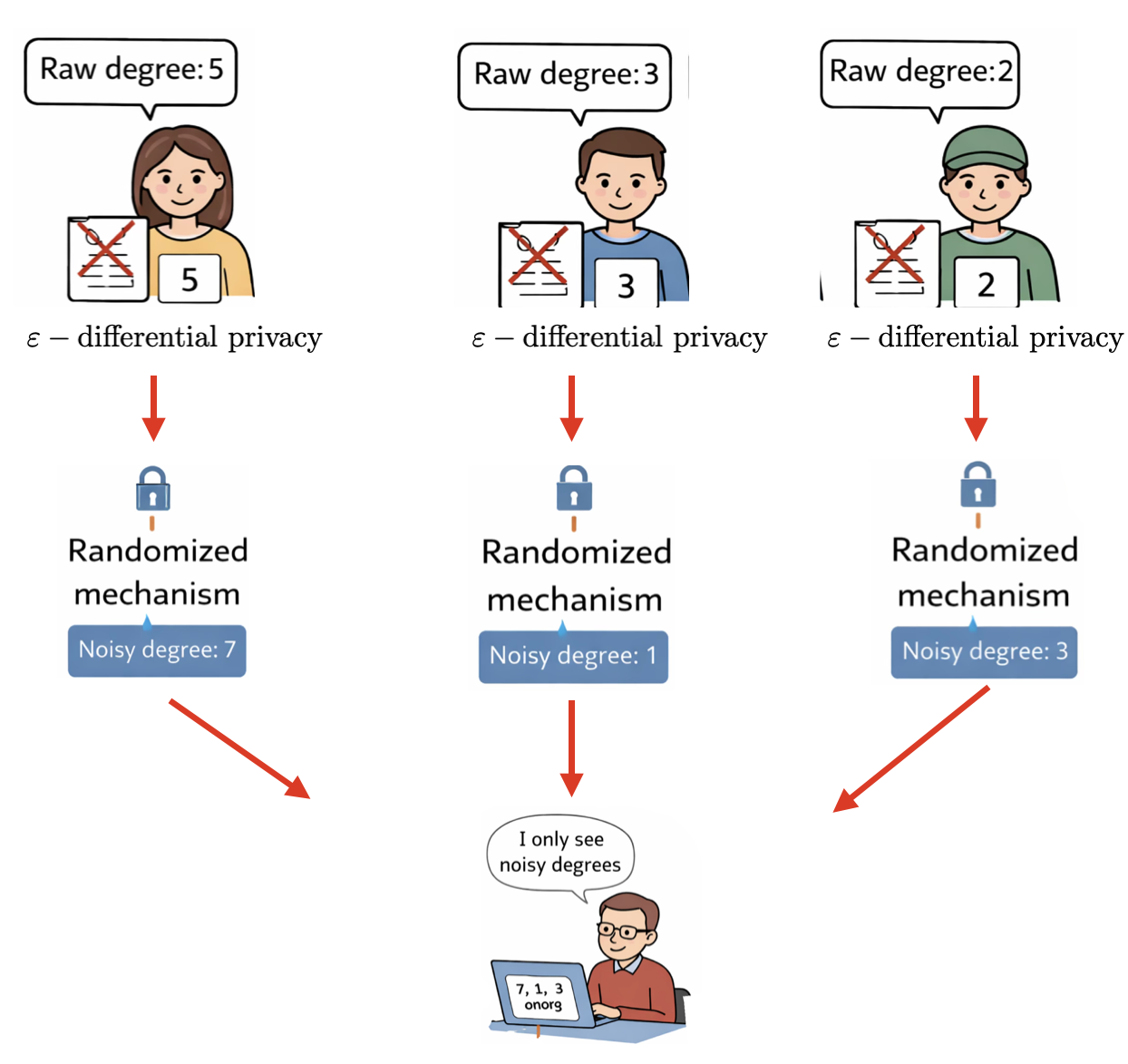}
    \caption{Local DP in hypergraph $\beta$-model}
  \end{minipage}
  \hfill
  \begin{minipage}[t]{0.48\linewidth}
    \centering
    \includegraphics[width=\linewidth]{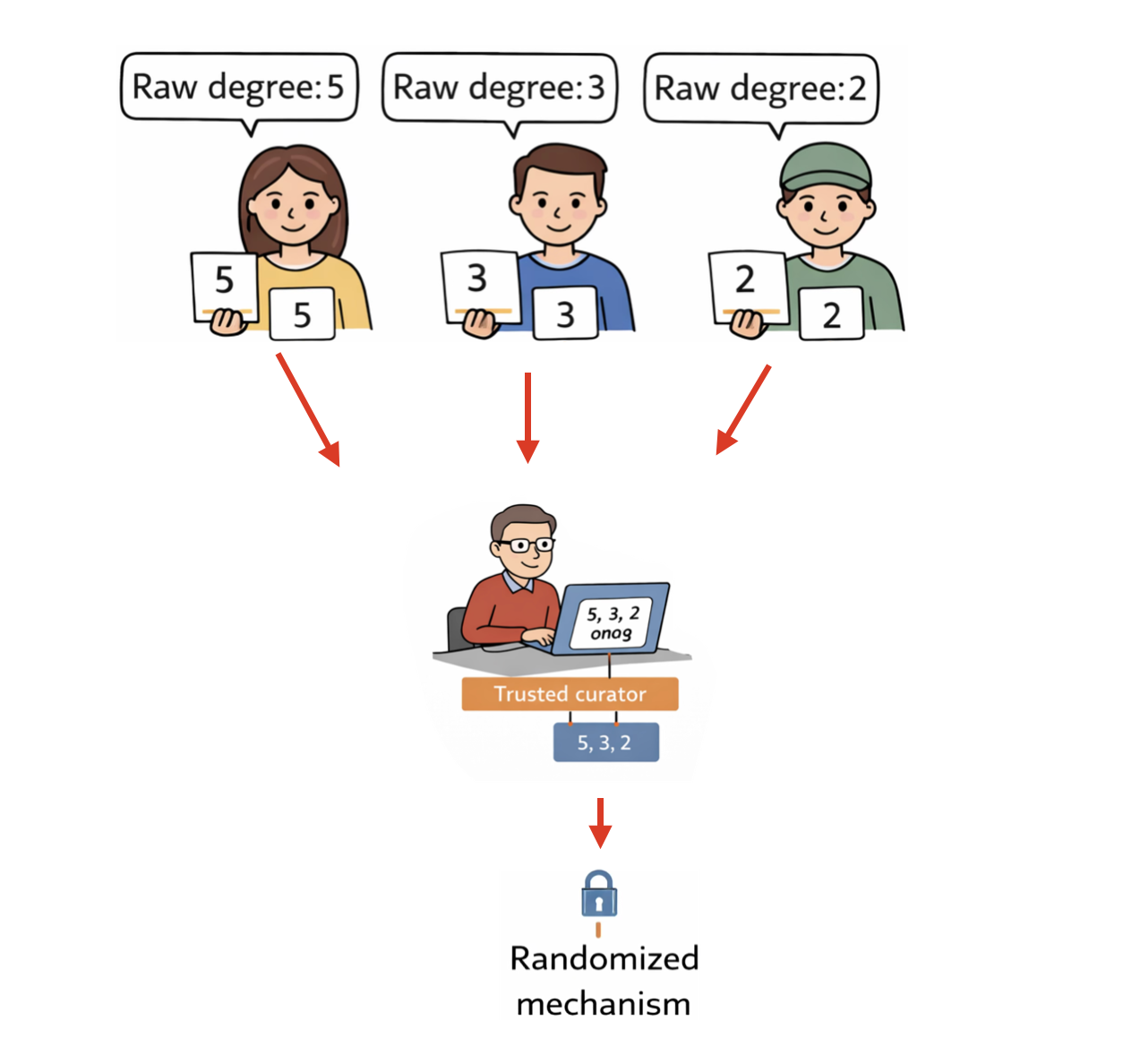}
    \caption{Central DP in hypergraph $\beta$-model}
  \end{minipage}
  \label{fig:dp_cartoon}
\end{figure}

\subsection{Privacy Framework}

In this paper, we adopt the \emph{edge differential privacy} framework introduced in \cite{nissim2007smooth}, where the goal is to protect the privacy of each individual (hyper-)edge against adversarial queries of the released information. We formalize this notion below.

\begin{definition}[Edge Differential Privacy]
A randomized mechanism $\mathcal M(\mathcal G)$ applied to information derived from a network $\mathcal G$ is said to satisfy $(\varepsilon,\delta)$-differential privacy if for any two fixed networks $\mathcal G$ and $\mathcal G'$ that differ in exactly one (hyper-)edge, and for any measurable set $\mathcal B$ in the output space of $\mathcal M$, the following holds:
\begin{align}
\mathbb P\!\left[\mathcal M(\mathcal G)\in\mathcal B\right]
\le
e^{\varepsilon}\mathbb P\!\left[\mathcal M(\mathcal G')\in\mathcal B\right]+\delta .
\end{align}
\end{definition}

We consider two privacy models within this framework: \emph{local differential privacy} and \emph{central differential privacy}.

\begin{enumerate}
\item {\bf Local Differential Privacy.}
In the local privacy model, each user perturbs their data before releasing it to the analyst. In our setting, this corresponds to perturbing the true degrees at the user level. The objective is to design a perturbation mechanism that ensures $\varepsilon$-differential privacy of the released degrees—and hence of any statistic computed from them—while incurring minimal loss in information about the model parameters.

\item {\bf Central Differential Privacy.}
In the central privacy model, users provide their data to a trusted curator, who releases a privatized version of aggregated statistics to the analyst. The goal in this framework is to design aggregation procedures that satisfy $(\varepsilon,\delta)$-differential privacy while preserving sufficient accuracy to enable meaningful statistical inference.
\end{enumerate}

\subsection{Major Contributions}
Our main contributions are summarized as follows.

\begin{enumerate}
\item {\bf Locally private release of degree information.}
Under the $\varepsilon$-local differential privacy framework, we show that a discrete Laplace mechanism obtained by adding carefully calibrated discrete Laplace noise  \citep{karwa_et_al} to the raw degrees is rate-optimal for parameter recovery while satisfying the prescribed privacy level. Building on minimax techniques for locally private estimation developed in \cite{duchi2017minimaxoptimalprocedureslocally}, we precisely characterize the worst-case privacy--utility trade-off as a function of the network size $n$ and the privacy parameter $\varepsilon$. To the best of our knowledge, such a sharp finite-sample characterization of the estimation risk in this setting has not appeared previously, even for the graph set-up.

\item {\bf Central privacy and DP gradient-based estimation.}
We also study a centrally private estimator of $\beta$ constructed via differentially private gradient descent-based optimization of the likelihood using the raw degrees. In this setting, we show that the resulting DP-GD-based estimator attains the minimax $\ell_2$ estimation risk among all $(\varepsilon,\delta)$-differentially private estimators, and we explicitly characterize the worst-case risk in terms of $n$ and $\varepsilon$. Our results further demonstrate that central differential privacy enables improved second-order error terms at the cost of weaker privacy guarantees.

\item {\bf Numerical studies and empirical validation.}
We complement our theoretical results with numerical experiments that illustrate the sharpness of the derived minimax bounds and compare the performance of locally and centrally private estimators across a range of privacy levels. We further validate our methods on a real communication network dataset, demonstrating the practical relevance of our privacy--utility characterizations.
\end{enumerate}

\subsection{Related Work}

The statistical analysis of $\beta$-models has a long and well-developed history. Early work on parameter estimation focused on accuracy guarantees for the maximum likelihood estimator, including $\ell_\infty$ error bounds \citep{2010arXiv1005.1136C} and characterization of conditions required for the existence of bounded MLE \citep{10.1214/12-AOS1078}. Several extensions of the $\beta$-model have since been studied, incorporating node covariates, temporal dependence, and other structural features of evolving networks \citep{Stein03042025,10.1214/17-AOS1585,10.1111/rssb.12444,https://doi.org/10.1111/sjos.12650}. More recently, \cite{nandy_bhattacharya} established minimax lower bounds for parameter estimation in hypergraph $\beta$-models under both $\ell_\infty$ and $\ell_2$ losses, as well as minimax detection thresholds for goodness-of-fit testing.

Related inferential problems for directed networks and pairwise comparison data have also been extensively investigated. These include $p_1$-models for directed random graphs and the Bradley--Terry--Luce (BTL) model for ranking from dyadic comparisons \citep{feng2025regressionanalysisreciprocitydirected,10.1093/biomet/asaf035,10.1093/imaiai/iaac032,Chen2021OptimalFR}. In particular, \cite{Cai12012026} derived sharp finite-sample minimax lower bounds for differentially private ranking under the BTL model.

In contrast, the literature on privacy-preserving network analysis has largely focused on community detection and clustering under differential privacy constraints \citep{mulle2015privacy,pinot2018graph,nguyen2016detecting,chakraborty2025primeprivacyawaremembershipprofile,wang2013learning,chen2023private}. Comparatively little attention has been paid to privacy--utility trade-offs for parameter estimation in degree-based network models. Existing work on differentially private estimation in the $\beta$-model is limited to asymptotic analyses for specific regimes under restrictive scalings of the privacy parameter $\varepsilon$ \citep{karwa_et_al} or access to the underlying network generating the observed degrees \citep{10.1214/24-AOS2365}. Furthermore, these works do not provide a precise finite-sample, minimax characterization of the estimation risk.

Our work seeks to fill this gap by providing the first finite-sample, minimax-optimal characterization of the privacy--utility trade-off for parameter estimation in graph and hypergraph $\beta$-models under both local and central differential privacy, together with matching upper and lower bounds that establish optimality of the proposed procedures.

\subsection{Notation and paper organization}
We adopt the following notation throughout the paper. The symbol $\mathbb{R}^k$ denotes the $k$-dimensional Euclidean space, and $\mathbb{N}$ denotes the set of natural numbers. For any $r \in \mathbb{N}$ and $x \in \mathbb{R}^k$, the $\ell_r$-norm is defined as $\|x\|_r = \left( \sum_{i=1}^k |x_i|^r \right)^{1/r}$, and the $\ell_\infty$-norm is defined as $\|x\|_\infty = \max_{1 \le i \le k} |x_i|$. For two nonnegative sequences $\{a_n\}$ and $\{b_n\}$, we write $a_n \lesssim_{\square} b_n$, and $a_n \asymp_{\square} b_n$ if there exist positive constants $C_1, C_2,$ and $C_3$, depending only on the quantities indicated in the subscript $\square$, such that $a_n \le C_1 b_n$ and $C_2 b_n \le a_n \le C_3 b_n$ for all $n \in \mathbb{N}$. When the subscript is omitted, the constants are absolute. Finally, for two sequences $\{a_n\}$ and $\{b_n\}$, we define $a_n=o(b_n)$, if and only if $a_n/b_n \rightarrow 0$, as $n \rightarrow \infty$.

The remainder of the paper is organized as follows. In Section~\ref{sec:privacy_utility_tradeoff}, we characterize the minimax rates for estimating the model parameters under both local and central differential privacy. In Section~\ref{sec:upper_bound_local}, we introduce a discrete Laplace mechanism for releasing the $r$-degrees of the observed network under $\varepsilon$-local differential privacy and present an estimator to achieve the corresponding minimax rate. In Section~\ref{sec:central_privacy}, we propose a DP--GD based estimator that satisfies $(\varepsilon,\delta)$-central differential privacy. In Section~\ref{sec:cost_param_est}, we empirically investigate the cost of privacy in parameter estimation under both privacy frameworks. Additional simulation results are deferred to the appendix. Finally, Section~\ref{sec:enron_email_dataset}, we present a real-data analysis of the Enron email dataset, comparing the cost of privacy in hyperlink prediction under local and central differential privacy using the proposed estimators. All proofs are deferred to the appendices.

\section{Privacy--Utility Trade-offs}
\label{sec:privacy_utility_tradeoff}

In this section, we provide an explicit characterization of the worst-case risk incurred by any private mechanism for parameter estimation in the hypergraph $\beta$-model. We adopt a minimax framework under the additional assumption that the true parameter vector satisfies $\|\beta\|_\infty \le M$. Such boundedness conditions are standard in the analysis of $\beta$-models and have been assumed, for example, in \cite{nandy_bhattacharya,Cai12012026}.

Let $\mathcal R_{\varepsilon,\delta,r}$ denote the class of estimators that satisfy $(\varepsilon,\delta)$-edge differential privacy in an $r$-uniform hypergraph $\beta$ model. We quantify the privacy--utility trade-off through a sequence $\{\varrho_{n,\varepsilon,\delta,r}\}$ defined as follows: there exists a constant $c_0(r,M)>0$ such that
\begin{align}
\label{eq:minimax_risk}
\min_{\wh \beta \in \mathcal R_{\varepsilon,\delta,r}}
\max_{\|\beta\|_\infty \le M}
\mathbb E\!\left[
\frac{1}{n}\|\wh \beta-\beta\|_2^2\right]
\ge c_0\, \varrho_{n,\varepsilon,\delta,r} .
\end{align}
The quantity $\varrho_{n,\varepsilon,\delta,r}$ thus represents the minimax estimation risk, up to constants, over all $(\varepsilon,\delta)$-differentially private estimators in the $r$-uniform hypergraph-$\beta$ model.

In the non-private setting, corresponding to $\delta=0$ and $\varepsilon \to \infty$, it is known from \cite{nandy_bhattacharya} that the minimax rate satisfies $\varrho_n \asymp n^{-(r-1)}$. Our goal is to characterize the additional statistical cost incurred by enforcing edge differential privacy at a prescribed $(\varepsilon,\delta)$ level, relative to this non-private benchmark.

\subsection{Lower bounds under local differential privacy}

Let us consider the class of estimators $\mathcal R^{\mathrm{loc}}_{\varepsilon,r}$ that are constructed from a released $r$-degree sequence in an $r$-uniform hypergraph, where the degrees are privatized locally by the nodes using a mechanism satisfying $\varepsilon$-\emph{edge local differential privacy}. Concretely, $\mathcal R^{\mathrm{loc}}_{\varepsilon,r}$ consists of estimators computed from the privatized degree sequence
\(
(\mathcal M(d_1),\ldots,\mathcal M(d_n)) \in \mathbb R^n,
\)
where $\mathcal M$ is a randomized mechanism satisfying
\begin{align}
\label{eq:def_local_priv_random}
\mathbb P\bigl[(\mathcal M(d_1),\ldots,\mathcal M(d_n))\in\mathcal B\bigr]
\;\le\;
e^{\varepsilon}
\mathbb P\bigl[(\mathcal M(d'_1),\ldots,\mathcal M(d'_n))\in\mathcal B\bigr]
\end{align}
for all Borel sets $\mathcal B$. Here $(d_1,\ldots,d_n)$ and $(d'_1,\ldots,d'_n)$ denote $r$-degree sequences arising from two $r$-uniform hypergraphs that differ in exactly one hyperedge. We denote by $\mathfrak M^{\mathrm{loc}}_{\varepsilon}$ the collection of all randomized mechanisms $\mathcal M$ satisfying \eqref{eq:def_local_priv_random}. We further restrict attention to estimators taking values in the bounded parameter space
\(
\Theta_M := \{\beta \in \mathbb R^n : \|\beta\|_\infty \le M\}.
\)
This restriction is without loss of generality, since any estimator can be clipped coordinatewise to $[-M,M]$ as a post-processing step, which preserves local differential privacy.

\begin{theorem}
\label{thm:low_bound_local}
For any $\varepsilon\in[0,\varepsilon_0]$, there exists a constant
$c_0=c_0(r,M,\varepsilon_0)>0$ such that
\begin{align}
\label{eq:minimax_risk_loc}
\min_{\widehat\beta \in \mathcal R^{\mathrm{loc}}_{\varepsilon,r}}\max_{\|\beta\|_\infty\le M}\mathbb E\!\left[\frac{1}{n}\|\widehat\beta-\beta\|_2^2\right]
\ge\frac{c_0}{\varepsilon^2 n^{r-1}}, \quad \mbox{for all $n \ge 16$.}
\end{align}
\end{theorem}

Theorem~\ref{thm:low_bound_local} shows that enforcing $\varepsilon$-local differential privacy increases the worst-case sample complexity required to achieve mean squared error at most $\eta>0$ by a factor of $O\left(\frac{1}{(\varepsilon^2)^{1/(r-1)}}\right)$.

The proof is based on a locally private version of Fano’s inequality \cite{duchi2017minimaxoptimalprocedureslocally}. We construct a packing of $2^{\Theta(n)}$ parameter vectors in $\mathbb R^n$ that are separated by order $\varepsilon^{-1}n^{-(r-1)/2}$ in $\ell_2$ norm, while the degree distributions induced by the corresponding $r$-uniform hypergraph $\beta$-models remain close in Kullback--Leibler divergence. This implies that, after local privatization, it is information-theoretically impossible to reliably distinguish among these candidate models. A key technical challenge is controlling the additional uncertainty introduced by the privatization mechanism; this requires sharper data-processing inequalities, which account for a further contraction of divergences by a factor of order $\varepsilon^2$ in the worst case. Full details are deferred to the appendix.

\subsection{Lower bounds under central differential privacy}

We next consider the class of estimators $\mathcal R^{\mathrm{cen}}_{\varepsilon,\delta,r}$ computed by a trusted curator who observes the true $r$-degree sequence from an $r$-uniform hypergraph generated according to the model \eqref{eq:r_uniform_beta_model}, and releases a privatized statistic satisfying $(\varepsilon,\delta)$-\emph{edge differential privacy}. Concretely, we consider randomized mechanisms $\mathcal M(d)$, where $d=(d_1,\ldots,d_n)\in\mathbb R^n$ denotes the observed $r$-degree sequence, such that
\[
\mathbb P\bigl[\mathcal M(d)\in\mathcal B\bigr]
\le
e^{\varepsilon}\mathbb P\bigl[\mathcal M(d')\in\mathcal B\bigr]
+\delta
\]
for all Borel sets $\mathcal B$. Here $d$ and $d'$ correspond to the $r$-degree sequences of two $r$-uniform hypergraphs differing in exactly one hyperedge. In this setting, the randomization is applied only to the aggregate statistic, rather than at the level of individual degrees. Furthermore, the estimators are also restricted to take values in $\Theta_M$. The resulting privacy--utility trade-off is characterized by the following theorem.

\begin{theorem}
\label{thm:low_bound_central}
Suppose $\varepsilon \in (0,\varepsilon_0]$ satisfies $\varepsilon n^{r-1} \ge c$ for some absolute constant $c>0$, and let $\delta =o(1)$. Then there exists a constant
$\mathfrak c_0:=\mathfrak c_0(r,M,\varepsilon_0,c,\delta)>0$ and some $n_0 \in \mathbb N$ such that
\begin{align}
\label{eq:minimax_risk_cen}
\min_{\widehat\beta \in \mathcal R^{\mathrm{cen}}_{\varepsilon,\delta,r}}
\max_{\|\beta\|_\infty\le M}
\mathbb E\!\left[\frac{1}{n}\|\widehat\beta-\beta\|_2^2\right]
\;\ge\;
\mathfrak c_0\cdot
\max\!\left\{
\frac{1}{n^{r-1}},\,
\frac{1}{\varepsilon^2 n^{2(r-1)}}
\right\}, \quad \mbox{for all $n \ge n_0$.}
\end{align}
\end{theorem}

In contrast to Theorem~\ref{thm:low_bound_local}, the privacy cost under central differential privacy appears only in the second-order term of the minimax risk, leading to strictly better estimation rates. This improvement, however, relies on the availability of a trusted curator with access to the raw degree sequence. The proof of Theorem~\ref{thm:low_bound_central} follows from a differentially private version of Assouad’s lemma developed in \cite{acharya2021dpassouad}.

\section{Locally Private Release of Degrees via the Discrete Laplace Mechanism}
\label{sec:upper_bound_local}

In this section, we describe a discrete Laplace mechanism–based protocol for
releasing the $r$-degree sequence under $\varepsilon$-local edge differential
privacy. The key idea is that each node independently perturbs its true
$r$-degree by adding calibrated discrete Laplace noise, so that the presence or
absence of any single hyperedge is statistically indistinguishable from the
random perturbation.

Let $G$ be a random $r$-uniform hypergraph generated from the $\beta$-model
\eqref{eq:r_uniform_beta_model}, and let
\(
d(G):=(d_1,\ldots,d_n)\in\mathbb R^n
\)
denote its $r$-degree sequence. We release a noisy version of the $r$-degree sequence by adding independent discrete Laplace noise to each coordinate. This guarantees that the effect of modifying a single hyperedge on the released sequence is obscured by the injected noise, thereby preventing an adversary from inferring the presence or absence of any particular hyperedge. We define the locally private degree sequence
$d^{\mathrm{loc}}(G)\in\mathbb R^n$ by 
\begin{align}
\label{eq:local_priv_deg_seq}
(d^{\mathrm{loc}}(G))_i=d_i + Z_i, \quad \mbox{for $i \in [n]$,}
\end{align}
where $Z_1,\ldots,Z_n \overset{\mathrm{i.i.d.}}{\sim}
\mathrm{Discrete\ Laplace}\!\left(e^{-\varepsilon/r}\right)$, with the
probability mass function given in Lemma~1 of \cite{karwa_et_al}.

\begin{theorem}
\label{thm:local_diff_priv}
The noisy degree sequence $d^{\mathrm{loc}}(G)$ satisfies $\varepsilon$-local edge
differential privacy.
\end{theorem}

\begin{remark}
The released sequence $d^{\mathrm{loc}}(G)$ need not correspond to a realizable
$r$-degree sequence of a hypergraph. In the
graph case ($r=2$), one may project noisy degree sequences onto the space of
graphical degree sequences \citep{karwa_et_al}. For hypergraphs, however, the
structure of the space of realizable $r$-degree sequences is substantially more
complex. While partial characterizations exist
\citep{10.1007/978-3-642-37067-0_26,FROSINI202197}, efficient projection procedures
are not currently available. Since our goal is inference on the model parameters rather than exact graph
reconstruction, we work directly with the perturbed degree sequence without
enforcing realizability.
\end{remark}
Given the locally private degree sequence, we define a regularized estimator
$\widehat\beta^{\mathrm{loc}}\in\mathbb R^n$ as
\begin{align}
\label{eq:local_mle}
\widehat\beta^{\mathrm{loc}}
=
\argmax_{\beta\in\mathbb R^n}
\left\{
\sum_{i=1}^n d^{\mathrm{loc}}_i \beta_i
-
\sum_{(i_1,\ldots,i_r)\in{[n]\choose r}}
\log(1+\exp(\beta_{i_1}+\cdots+\beta_{i_r}))
-
\lambda \|\beta\|_2^2
\right\},
\end{align}
where $\lambda>0$ is a regularization parameter. 
Since $\wh \beta^{\mathrm{loc}}$ is computed through a post-processing operation applied
to an $\varepsilon$-locally private degree sequence, the resulting estimator
$\widehat\beta^{\mathrm{loc}}$ remains $\varepsilon$-local edge differentially
private. The convergence rate of $\widehat\beta^{\mathrm{loc}}$ is given in the following theorem.

\begin{theorem}
\label{thm:local_conv_rate}
Assume that $\lambda \asymp n^{-(r-1)/2}$, and that there exists a solution
$\widehat\beta^{\mathrm{loc}} \in \mathcal B_{\infty,M}$ to \eqref{eq:local_mle}. Then there exists a constant
$C_0=C_0(r,M)>0$ such that, for all $\beta\in\mathcal B_{\infty,M}$,
\[
\frac{1}{n}\mathbb E\!\left[
\|\widehat\beta^{\mathrm{loc}}-\beta\|_2^2
\right]
\le
C_0\,\frac{\log n}{n^{r-1}\varepsilon^2}.
\]
\end{theorem}
The above theorem shows that the discrete Laplace mechanism-based release of the $r$-degree sequence allows for estimation of $\beta$ in \eqref{eq:r_uniform_beta_model} with the minimax convergence rate (up to logarithmic terms) with a edge differentially private estimator.

\begin{remark}
The assumption that a bounded solution to the likelihood equations exists is essential for the conclusion of Theorem~\ref{thm:local_conv_rate}. For $r=2$, this issue can be addressed by projecting the noisy degree sequence back onto the space of graphical degree sequences, as proposed in \cite{karwa_et_al}, which guarantees the existence of a bounded maximum likelihood estimator provided the original degree sequence satisfied some conditions. For $r\ge 3$, however, the structure of hypergraph degree sequences is substantially more complex, and analogous projection procedures are not currently available. While one may enforce boundedness through stronger regularization, such modifications typically introduce additional bias and lead to estimators that are no longer rate optimal. At present, it remains unclear whether bounded solutions to the likelihood equations exist under minimal conditions for $r\ge 3$. We therefore view this assumption as a technical but unavoidable requirement for establishing sharp convergence rates. More broadly, we conjecture that in higher-order settings, the potential nonexistence of bounded maximum likelihood solutions may reflect an intrinsic gap between information-theoretic lower bounds and the performance of computationally tractable estimators.
\end{remark}

\section{DP-GD Based Parameter Estimation under Central Differential Privacy}
\label{sec:central_privacy}

In this section, we develop a differentially private gradient descent (DP-GD)
framework for estimating the parameter vector $\beta$ from the true
$r$-degree sequence $d=(d_1,\ldots,d_n)$ under the central differential privacy
model. In this setting, a trusted curator has access to the true degree sequence
and computes a regularized maximum likelihood estimator using gradient descent,
while injecting carefully calibrated Gaussian noise into the gradients at each
iteration to ensure $(\varepsilon,\delta)$-edge differential privacy.

Specifically, consider the regularized negative log-likelihood function
\begin{align}
\ell_n(\beta)
:=
\frac{1}{{n\choose r}}
\left\{
\sum_{(i_1,\ldots,i_r)\in{[n]\choose r}}
\log(1+\exp(\beta_{i_1}+\cdots+\beta_{i_r}))-\sum_{i=1}^n d_i \beta_i
\right\}.
\end{align}
We maximize $\ell_n(\beta)$ using projected gradient descent with additive
Gaussian noise. Observe that for any $\gamma\in\mathbb R^n$ satisfying
$\|\gamma\|_\infty\le M$, the Hessian of $\ell_n(\gamma)$ satisfies
\begin{align}
\label{eq:get_strong_convexity_smooth}
\tfrac1{4n} e^{-2rM}\;\le\;\lambda_{\min}(\nabla^2\ell_n(\gamma))\;\le\;\lambda_{\max}(\nabla^2\ell_n(\gamma))\;\le\;\frac{1}{n}.
\end{align}
We initialize $\beta^{(0)}=0$, set the step size
$\eta=0.25\,ne^{-2rM}$, and run the algorithm for
\(
T = 32(r-1)e^{4rM}\{(r-1)\log(n)+2\log M\},
\)
many iterations.
For $t=1,\ldots,T$, the update is given by
\begin{align}
\label{eq:dp_gd_updates}
\beta^{(t+1)}
:=
\mathsf{Proj}_{\mathcal B_{\infty,M}}
\Bigl(
\beta^{(t)} - \eta\bigl[\nabla\ell_n(\beta^{(t)}) + Z_t\bigr]
\Bigr),
\end{align}
where
\[
Z_1,\ldots,Z_T
\;\overset{\mathrm{i.i.d.}}{\sim}\;
\mathcal N_n\!\left(
0,\,
4rT\,n^{-2r}\varepsilon^{-2}\log(1/\delta)\,I_n
\right).
\]
We define the final estimator as $\widehat\beta^{\mathrm{cen}}:=\beta^{(T)}$. An important feature of the algorithm is a parameter dependent early stopping of the iterations which prevents excessive noise injection into the optimization process while ensuring privacy at the user specified level. 

\begin{theorem}
\label{thm:central_privacy_valid}
The estimator $\widehat\beta^{\mathrm{cen}}$ satisfies $(\varepsilon,\delta)$-edge
differential privacy.
\end{theorem}
We next characterize the statistical accuracy of
$\widehat\beta^{\mathrm{cen}}$.
\begin{theorem}
\label{thm:bound_central_privacy}
Let $\varepsilon\in(0,\varepsilon_0]$, and $\delta\in(0,1)$. Suppose there exists a maximizer $\widehat\beta \in \mathcal B_{\infty,M}$ of the objective function $\ell_n(\beta)$. Then there exists a
constant $\mathfrak C_0:=\mathfrak C_0(r,M,\varepsilon_0)>0$ such that
\begin{align}
\frac{1}{n}\mathbb E\!\left[\|\widehat\beta^{\mathrm{cen}}-\beta\|_2^2\right]
\;\le\;
\mathfrak C_0\cdot\max\!\left\{\frac{1}{n^{r-1}},\frac{(\log n)\log(1/\delta)}{n^{2r-2}\varepsilon^2}
\right\}.
\end{align}
\end{theorem}
Theorem~\ref{thm:bound_central_privacy} shows that the DP-GD estimator
$\widehat\beta^{\mathrm{cen}}$ achieves the optimal dependence on $\varepsilon$
and the correct polynomial scaling in the network size $n$ among
$(\varepsilon,\delta)$-edge differentially private estimators (Theorem \ref{thm:low_bound_central}), up to logarithmic
factors in $n$ and $\delta$. 

\section{Numerical Experiment: Cost of privacy in parameter estimation}
\label{sec:cost_param_est}
In this section, we empirically study the performance of the estimators $\wh \beta^{\mathrm{loc}}$ (defined in \eqref{eq:local_mle}) and $\wh \beta^{\mathrm{cen}}$ (defined in \eqref{eq:dp_gd_updates}) as functions of the privacy budget $\varepsilon$ and the network size $n$. We emphasize that our contribution is \textbf{theoretical rather than methodological}: the primary goal is to characterize the worst-case statistical cost of enforcing differential privacy. As established in Theorems~\ref{thm:local_conv_rate} and~\ref{thm:bound_central_privacy}, the impact of privacy is most significant when the network size is relatively small. Accordingly, we restrict attention to $3$-uniform hypergraphs with $n \in \{50,100,150,200\}$. We consider privacy budgets $\varepsilon \in \{0.0001, 0.001, 0.01, 0.1, 1\}$ and evaluate the performance of the foregoing methods in terms of parameter estimation for the $3$-uniform hypergraph $\beta$-model.

\begin{figure}[t]
  \centering
  \begin{minipage}[t]{0.48\linewidth}
    \centering
    \includegraphics[width=\linewidth]{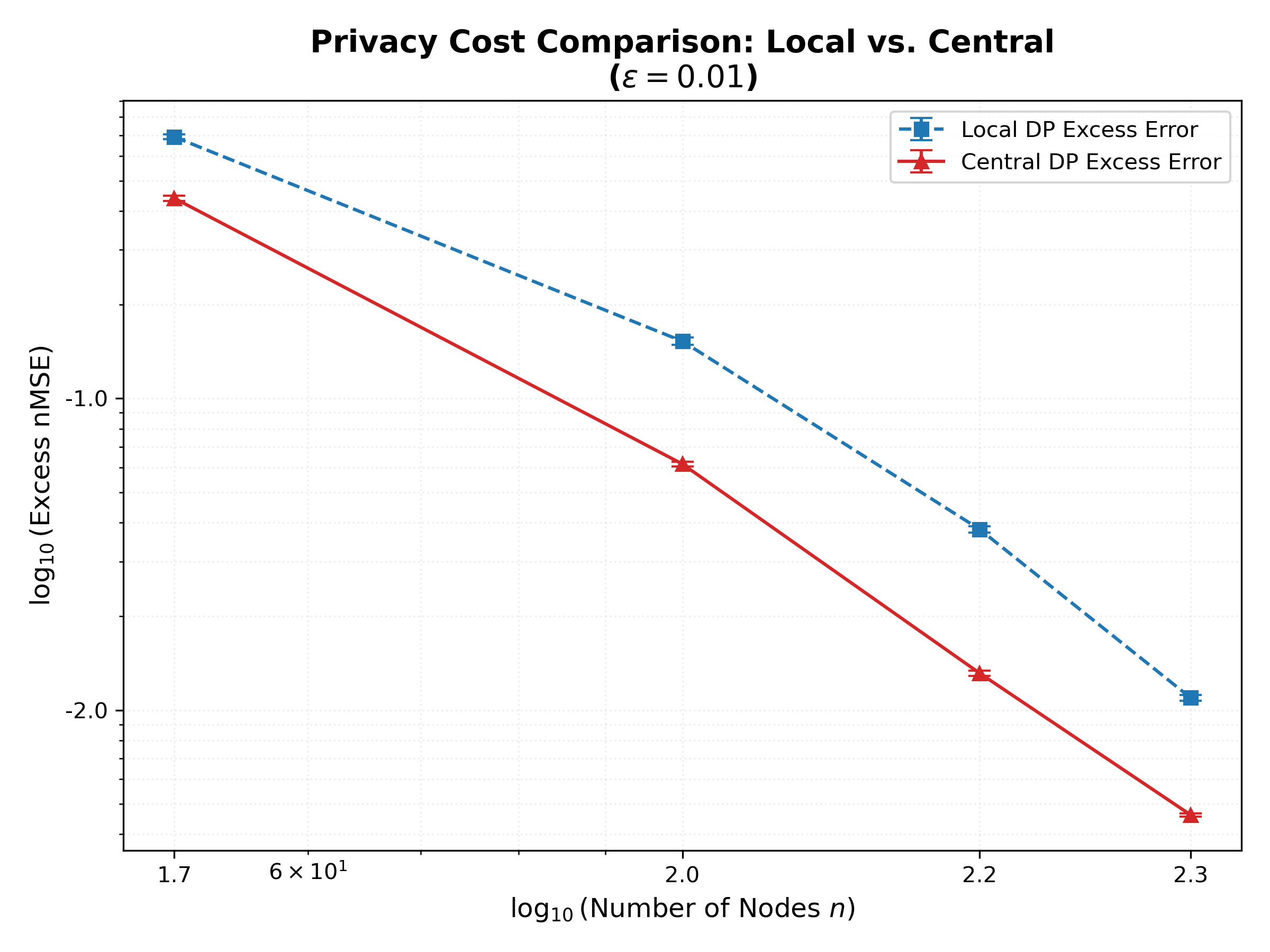}
    \caption{Log excess risk as a function of $\log_{10} n$ for $\varepsilon=0.01$.}
    \label{fig:vary_n}
  \end{minipage}
  \hfill
  \begin{minipage}[t]{0.48\linewidth}
    \centering
    \includegraphics[width=\linewidth]{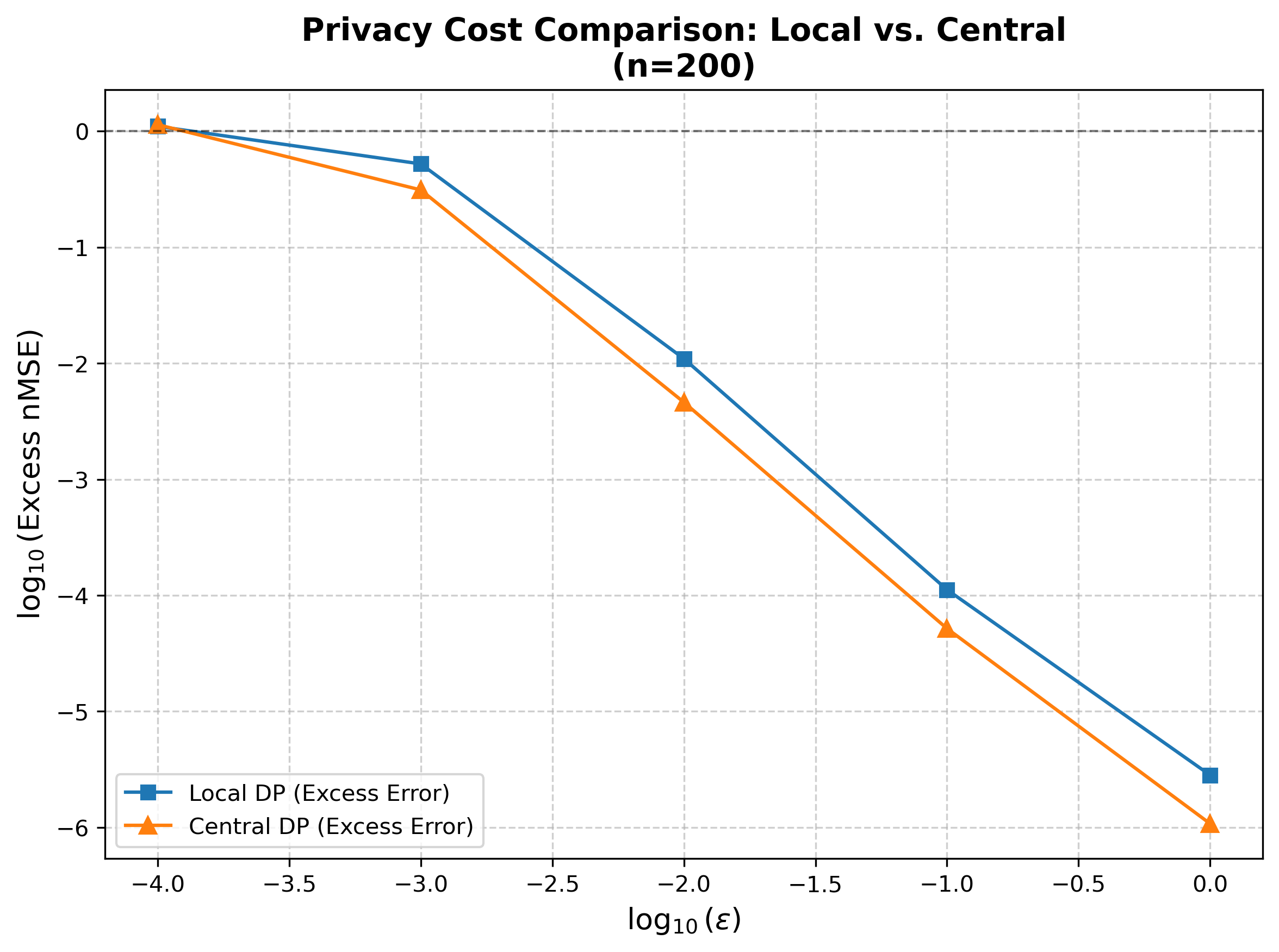}
    \caption{Log excess risk as a function of $\log_{10}\varepsilon$ for $n=200$.}
    \label{fig:vary_eps}
  \end{minipage}
\end{figure}

We generate the true parameters $\beta\in \mathbb R^n$ according to
\[
\beta_i \overset{\mathrm{i.i.d.}}{\sim} 0.3\,\delta_0 + 0.7\,\mathcal N_{[-M,M]}(\mu,\sigma^2),
\]
where $\mathcal N_{[-M,M]}(\mu,\sigma^2)$ denotes a normal distribution with mean $\mu$ and variance $\sigma^2$ truncated to $[-M,M]$. Throughout this experiment, we set $\mu=0.5$, $\sigma=0.02$, and $M=1$. The estimators $\wh \beta^{\mathrm{loc}}$ and $\wh \beta^{\mathrm{cen}}$ are computed using gradient descent applied to \eqref{eq:local_mle} and the DP-GD updates in \eqref{eq:dp_gd_updates}, respectively, each run for $T=10{,}000$ iterations with a common step size $\eta := 0.25\,n e^{-2rM}$. For numerical stability, the objective in \eqref{eq:local_mle} is normalized by ${n \choose 3}$.

Each configuration is repeated independently $50$ times, and we report the average normalized mean squared error (nMSE). We define the excess risks as
\begin{align}
\text{Excess $\mathrm{nMSE}_{\mathrm{loc}}$}
&:= \frac{1}{n}\|\wh \beta^{\mathrm{loc}}-\beta\|_2^2
 - \frac{1}{n}\|\wh \beta_{\mathrm{MLE},\lambda}-\beta\|_2^2, \nonumber\\
\text{Excess $\mathrm{nMSE}_{\mathrm{cen}}$}
&:= \frac{1}{n}\|\wh \beta^{\mathrm{cen}}-\beta\|_2^2
 - \frac{1}{n}\|\wh \beta_{\mathrm{MLE}}-\beta\|_2^2 .
\end{align}
Here, $\wh \beta_{\mathrm{MLE},\lambda}$ denotes the regularized MLE obtained by optimizing \eqref{eq:local_mle} using unperturbed degrees with $\lambda=0.0001\,n$, which serves as minimal regularization to stabilize the updates, while $\wh \beta_{\mathrm{MLE}}$ is the unregularized MLE. The latter is used as the baseline for the central DP estimator to maintain parity with the DP-GD updates. Although it is possible to incorporate regularization into \eqref{eq:dp_gd_updates}, its effect is negligible for the chosen value of $\lambda$. For central DP, we set $\delta=n^{-2}$.

Figure~\ref{fig:vary_n} shows that the logarithm of excess $\mathrm{nMSE}$ decreases as $\log_{10} n$ increases, with an approximately linear decay consistent with the rates predicted in Theorems~\ref{thm:local_conv_rate} and~\ref{thm:bound_central_privacy}. Moreover, the decay is faster for the central DP estimator than for the local DP estimator, reflecting the weaker statistical cost of privacy under central DP, albeit at the expense of stronger trust assumptions. A similar phenomenon is observed in Figure~\ref{fig:vary_eps}: in the central privacy setting, the log–log plot (base 10) follows the same slope predicted by our theory, but with a noticeably smaller intercept. Overall, these figures suggest that the dependence of the excess risk on $\varepsilon$ is substantially weaker for the central DP estimator than for its local DP counterpart.\footnote{An additional numerical experiment studying the cost of privacy in link-prediction in a $3$-uniform hypergraph is provided in the appendix.}

\begin{table*}[t]
\centering
\scriptsize
\caption{Comparison of private and non-private estimators for $3$-uniform hypergraph link prediction.}
\label{tab:results_enron_link_pred}
\begin{tabular}{llccc}
\toprule
\textbf{$\varepsilon$} & \textbf{Method} & \textbf{ROC-AUC} & \textbf{AP} & \textbf{ECE} \\
\midrule
--- & MLE (non-private)        & 0.817 & 0.793 & 0.271 \\
--- & Ridge MLE (non-private)  & 0.817 & 0.793 & 0.271 \\
\midrule
\multirow{2}{*}{$0.001$}
  & Central DP-GD & 0.521 & 0.537 & 0.281 \\
  & Local DP      & 0.430 & 0.479 & 0.240 \\
\cmidrule(lr){1-5}
\multirow{2}{*}{$0.01$}
  & Central DP-GD & 0.535 & 0.546 & 0.273 \\
  & Local DP      & 0.434 & 0.481 & 0.269 \\
\cmidrule(lr){1-5}
\multirow{2}{*}{$0.1$}
  & Central DP-GD & 0.658 & 0.678 & 0.271 \\
  & Local DP      & 0.609 & 0.636 & 0.271 \\
\cmidrule(lr){1-5}
\multirow{2}{*}{$1.0$}
  & Central DP-GD & 0.817 & 0.791 & 0.271 \\
  & Local DP      & 0.803 & 0.784 & 0.271 \\
\bottomrule
\end{tabular}
\end{table*}

\section{Privacy-preserving link prediction on the Enron email dataset}
\label{sec:enron_email_dataset}
We study privacy-preserving link prediction on the Enron email dataset \citep{Benson-2018-simplicial}, which records email communications among employees of the Enron corporation. The data can be represented as a hypergraph on $143$ employees, where each hyperedge corresponds to a group of individuals involved in a single email. The original dataset contains $10{,}883$ time-stamped hyperedges. To remove duplication, we retain only the first occurrence of each unique group and restrict attention to hyperedges of size three, yielding a final dataset of $317$ distinct $3$-uniform hyperedges on $125$ nodes.

For link prediction, we randomly hold out $20\%$ of the observed hyperedges and independently sample an equal number of non-edges to form a balanced test set of $64$ candidate hyperlinks. The remaining hyperedges are used for training, and parameter estimation is based solely on the degree sequence computed from the training data. We compute four estimators: the non-private maximum likelihood estimator $\widehat\beta_{\mathrm{MLE}}$, its ridge-regularized version $\widehat\beta_{\mathrm{MLE},\lambda}$, the centrally private estimator $\widehat\beta^{\mathrm{cen}}$, and the locally private estimator $\widehat\beta^{\mathrm{loc}}$. All estimators are obtained via gradient-based optimization. Since the true bound $M$ on the parameter vector is unknown for this real dataset, we adopt the conservative choice $M = 2\sqrt{\log n}$ and use a fixed step size $\eta = 0.005$, which is slightly more aggressive than the worst-case theoretical step size but empirically stable. Each estimator is computed using $T = 10{,}000$ gradient descent iterations. While imperfect tuning of these hyperparameters can degrade private estimators, our choices are intentionally conservative and avoid artificially inflating performance; in practice, such tuning is often unavoidable and problem-specific. For the private estimators, we vary the privacy budget over $\varepsilon \in \{0.001, 0.01, 0.1, 1\}$ and fix $\delta = n^{-2}$. Using the fitted parameters, we compute link probabilities for candidate test hyperedges according to the $3$-uniform $\beta$-model in \eqref{eq:r_uniform_beta_model}. Predictive performance is evaluated using ROC--AUC \citep{hanley1982meaning} and average precision (AP) \citep{vanRijsbergen1979IR}, while probability calibration is assessed using the expected calibration error (ECE) \citep{10.5555/3305381.3305518} \footnote{For details on these evaluation metrics, see Section \ref{sec:pred_mertics_def} in the appendix}. Results are summarized in Table~\ref{tab:results_enron_link_pred}. Both ROC--AUC and AP reveal a clear cost of privacy, with substantially larger degradation under local privacy than under central privacy, and a monotone improvement as $\varepsilon$ increases. Across all methods, calibration remains poor, reflecting the inherent difficulty of probability estimation in extremely sparse hypergraph settings. Overall, these findings are consistent with our theoretical results, which predict a higher cost under local differential privacy and improved utility with increasing privacy budget.

\section{Discussion and future directions}

In this paper, we characterize the fundamental information-theoretic costs of parameter estimation in higher-order networks with pronounced degree heterogeneity under differential privacy constraints on individual hyperlinks. We study both central and local differential privacy frameworks, establish minimax estimation rates, and provide computationally efficient algorithms that achieve these rates up to constants and logarithmic factors. While our analysis focuses on the unregularized MLE in the central-DP setting, analogous rates can be derived for regularized estimators; with appropriate regularization, the dependence on $n$ and $\varepsilon$ deteriorates at most by logarithmic factors.

Several important questions remain open. A primary challenge is to characterize conditions on the hyper-degree sequence under which the MLE admits a bounded solution. Even in the non-private regime, a complete understanding of this issue is lacking for $r \ge 3$. A natural conjecture is that when the MLE fails to exist, a gap emerges between the information-theoretic lower bound on estimation risk and what is achievable by polynomial-time algorithms.

Beyond these questions, the $\beta$-model itself is highly stylized. A more realistic alternative is provided by latent space models, where high-dimensional latent vectors govern hyperlink formation. Characterizing the cost of privacy in such models is substantially more challenging and would require nontrivial extensions of the present theory. Another natural direction is to study node-level differential privacy in the central DP framework, ensuring that the release of parameters does not reveal the presence or absence of any individual node. Finally, it would be interesting to move beyond linear log-odds models to nonlinear parameterizations, for instance via neural networks, and to explore whether flow-based or transfer-learning methods can leverage information from large, dense networks to improve estimation in smaller or sparser settings. These directions remain promising avenues for future research.

\paragraph{Use of generative AI.}
The authors employed ChatGPT 5.2 and Gemini for text editing and coding assistance. The core scientific contributions—including proof concepts, methodology, and experimental design—were developed solely by the authors.
\section*{Acknowledgement}
We thank Bhaswar B. Bhattacharya for introducing us to the problems and many helpful suggestions. Furthermore, SN thanks Abhinav Chakraborty for many useful discussions that helped us prove the lower bounds.

\bibliographystyle{apalike}
\bibliography{private_beta_model}

@article{chen2014correlated,
  title={Correlated network data publication via differential privacy},
  author={Chen, Rui and Fung, Benjamin CM and Yu, Philip S and Desai, Bipin C},
  journal={The VLDB Journal},
  volume={23},
  number={4},
  pages={653--676},
  year={2014},
  publisher={Springer}
}

@article{duchi2017minimaxoptimalprocedureslocally,
      title={Minimax Optimal Procedures for Locally Private Estimation}, 
      author={John Duchi and Martin Wainwright and Michael Jordan},
      year={2017},
      journal={arxiv preprint: 1604.02390},
      archivePrefix={arXiv},
      primaryClass={math.ST},
      url={https://arxiv.org/abs/1604.02390}, 
}

@article{park2004statistical,
  title={Statistical mechanics of networks},
  author={Park, Juyong and Newman, Mark EJ},
  journal={Physical Review E—Statistical, Nonlinear, and Soft Matter Physics},
  volume={70},
  number={6},
  pages={066117},
  year={2004},
  publisher={APS}
}

@ARTICLE{2010arXiv1005.1136C,
author = {Sourav Chatterjee and Persi Diaconis and Allan Sly},
title = {{Random graphs with a given degree sequence}},
volume = {21},
journal = {The Annals of Applied Probability},
number = {4},
publisher = {Institute of Mathematical Statistics},
pages = {1400 -- 1435},
year = {2011},
doi = {10.1214/10-AAP728},
URL = {https://doi.org/10.1214/10-AAP728}
}

@ARTICLE{nandy_bhattacharya,
  author={Nandy, Sagnik and Bhattacharya, Bhaswar B.},
  journal={IEEE Transactions on Information Theory}, 
  title={Degree Heterogeneity in Higher-Order Networks: Inference in the Hypergraph $\beta$-Model}, 
  year={2024},
  volume={70},
  number={8},
  pages={6000-6024},
  doi={10.1109/TIT.2024.3411523}}

@inproceedings{nguyen2015differentially,
  title={Differentially private publication of social graphs at linear cost},
  author={Nguyen, Hiep H and Imine, Abdessamad and Rusinowitch, Micha{\"e}l},
  booktitle={Proceedings of the 2015 IEEE/ACM International Conference on Advances in Social Networks Analysis and Mining 2015},
  pages={596--599},
  year={2015}
}

@article{dwork2014algorithmic,
  title={The algorithmic foundations of differential privacy},
  author={Dwork, Cynthia and Roth, Aaron and others},
  journal={Foundations and trends{\textregistered} in theoretical computer science},
  volume={9},
  number={3--4},
  pages={211--407},
  year={2014},
  publisher={Now Publishers, Inc.}
}

@inproceedings{mulle2015privacy,
  title={Privacy-Integrated Graph Clustering Through Differential Privacy.},
  author={M{\"u}lle, Yvonne and Clifton, Chris and B{\"o}hm, Klemens},
  booktitle={EDBT/ICDT Workshops},
  volume={1330},
  pages={247--254},
  year={2015}
}

@article{helleringer2007sexual,
  title={Sexual network structure and the spread of HIV in Africa: evidence from Likoma Island, Malawi},
  author={Helleringer, Stephane and Kohler, Hans-Peter},
  journal={Aids},
  volume={21},
  number={17},
  pages={2323--2332},
  year={2007},
  publisher={LWW}
}

@inproceedings{nissim2007smooth,
  title={Smooth sensitivity and sampling in private data analysis},
  author={Nissim, Kobbi and Raskhodnikova, Sofya and Smith, Adam},
  booktitle={Proceedings of the thirty-ninth annual ACM symposium on Theory of computing},
  pages={75--84},
  year={2007}
}

@inproceedings{kasiviswanathan2013analyzing,
  title={Analyzing graphs with node differential privacy},
  author={Kasiviswanathan, Shiva Prasad and Nissim, Kobbi and Raskhodnikova, Sofya and Smith, Adam},
  booktitle={Theory of Cryptography Conference},
  pages={457--476},
  year={2013},
  organization={Springer}
}

@article{hay2007anonymizing,
  title={Anonymizing social networks},
  author={Hay, Michael and Miklau, Gerome and Jensen, David and Weis, Philipp and Srivastava, Siddharth},
  year={2007},
  publisher={Technical Report 07-19, University of Massachusetts Amherst}
}

@inproceedings{narayanan2009anonymizing,
  title={De-anonymizing social networks},
  author={Narayanan, Arvind and Shmatikov, Vitaly},
  booktitle={2009 30th IEEE symposium on security and privacy},
  pages={173--187},
  year={2009},
  organization={IEEE}
}

@article{10.1214/12-AOS1078,
author = {Alessandro Rinaldo and Sonja Petrović and Stephen E. Fienberg},
title = {{Maximum lilkelihood estimation in the $\beta$-model}},
volume = {41},
journal = {The Annals of Statistics},
number = {3},
publisher = {Institute of Mathematical Statistics},
pages = {1085 -- 1110},
year = {2013},
doi = {10.1214/12-AOS1078},
URL = {https://doi.org/10.1214/12-AOS1078}
}

@article{Stein03042025,
author = {Stefan Stein and Rui Feng and Chenlei Leng},
title = {A Sparse Beta Regression Model for Network Analysis},
journal = {Journal of the American Statistical Association},
volume = {120},
number = {550},
pages = {1281--1293},
year = {2025},
publisher = {Taylor \& Francis},
doi = {10.1080/01621459.2024.2411073},
URL = {    
        https://doi.org/10.1080/01621459.2024.2411073
},
eprint = {  
        https://doi.org/10.1080/01621459.2024.2411073
}
}

@article{Chen2021OptimalFR,
  title={Optimal full ranking from pairwise comparisons},
  author={Pinhan Chen and Chao Gao and Anderson Y. Zhang},
  journal={The Annals of Statistics},
  year={2021},
  url={https://api.semanticscholar.org/CorpusID:231662262}
}

@inproceedings{abadi_et_al,
author = {Abadi, Martin and Chu, Andy and Goodfellow, Ian and McMahan, H. Brendan and Mironov, Ilya and Talwar, Kunal and Zhang, Li},
title = {Deep Learning with Differential Privacy},
year = {2016},
isbn = {9781450341394},
publisher = {Association for Computing Machinery},
address = {New York, NY, USA},
url = {https://doi.org/10.1145/2976749.2978318},
doi = {10.1145/2976749.2978318},
booktitle = {Proceedings of the 2016 ACM SIGSAC Conference on Computer and Communications Security},
pages = {308–318},
numpages = {11},
location = {Vienna, Austria},
series = {CCS '16}
}

@article{Cai12012026,
author = {T. Tony Cai and Abhinav Chakraborty and Yichen Wang},
title = {Optimal Differentially Private Ranking from Pairwise Comparisons*},
journal = {Journal of the American Statistical Association},
volume = {0},
number = {ja},
pages = {1--25},
year = {2026},
publisher = {Taylor \& Francis},
doi = {10.1080/01621459.2026.2612773},
URL = {   
        https://doi.org/10.1080/01621459.2026.2612773
},
eprint = { 
        https://doi.org/10.1080/01621459.2026.2612773
}}

@article{10.1093/imaiai/iaac032,
    author = {Gao, Chao and Shen, Yandi and Zhang, Anderson Y},
    title = {Uncertainty quantification in the Bradley–-Terry-–Luce model},
    journal = {Information and Inference: A Journal of the IMA},
    volume = {12},
    number = {2},
    pages = {1073-1140},
    year = {2023},
    month = {01},
    issn = {2049-8772},
    doi = {10.1093/imaiai/iaac032},
    url = {https://doi.org/10.1093/imaiai/iaac032},
}

@article{feng2025regressionanalysisreciprocitydirected,
      title={Regression Analysis of Reciprocity in Directed Networks}, 
      author={Rui Feng and Chenlei Leng},
      year={2025},
      journal={arxiv preprint: 2507.21469},
      archivePrefix={arXiv},
      primaryClass={stat.ME},
      url={https://arxiv.org/abs/2507.21469}, 
}

@article{10.1093/biomet/asaf035,
    author = {Feng, Rui and Leng, Chenlei},
    title = {Modelling directed networks with reciprocity},
    journal = {Biometrika},
    volume = {112},
    number = {2},
    pages = {asaf035},
    year = {2025},
    month = {06},
    issn = {1464-3510},
    doi = {10.1093/biomet/asaf035},
    url = {https://doi.org/10.1093/biomet/asaf035},
    eprint = {https://academic.oup.com/biomet/article-pdf/112/2/asaf035/63417429/asaf035.pdf},
}

@article{https://doi.org/10.1111/sjos.12650,
author = {Du, Yuqing and Qu, Lianqiang and Yan, Ting and Zhang, Yuan},
title = {Time-varying $\beta$-model for dynamic directed networks},
journal = {Scandinavian Journal of Statistics},
volume = {50},
number = {4},
pages = {1687-1715},
doi = {https://doi.org/10.1111/sjos.12650},
url = {https://onlinelibrary.wiley.com/doi/abs/10.1111/sjos.12650},
eprint = {https://onlinelibrary.wiley.com/doi/pdf/10.1111/sjos.12650},
year = {2023}
}

@article{10.1111/rssb.12444,
    author = {Chen, Mingli and Kato, Kengo and Leng, Chenlei},
    title = {Analysis of Networks via the Sparse $\beta$-model},
    journal = {Journal of the Royal Statistical Society Series B: Statistical Methodology},
    volume = {83},
    number = {5},
    pages = {887-910},
    year = {2021},
    month = {09},
    issn = {1369-7412},
    doi = {10.1111/rssb.12444},
    url = {https://doi.org/10.1111/rssb.12444},
    eprint = {https://academic.oup.com/jrsssb/article-pdf/83/5/887/49322040/jrsssb_83_5_887.pdf},
}

@article{10.1214/24-AOS2365,
author = {Jinyuan Chang and Qiao Hu and Eric D. Kolaczyk and Qiwei Yao and Fengting Yi},
title = {{Edge differentially private estimation in the $\beta$-model via jittering and method of moments}},
volume = {52},
journal = {The Annals of Statistics},
number = {2},
publisher = {Institute of Mathematical Statistics},
pages = {708 -- 728},
year = {2024},
doi = {10.1214/24-AOS2365},
URL = {https://doi.org/10.1214/24-AOS2365}
}

@article{holland1981exponential,
  title={An exponential family of probability distributions for directed graphs},
  author={Holland, Paul W and Leinhardt, Samuel},
  journal={Journal of the american Statistical association},
  volume={76},
  number={373},
  pages={33--50},
  year={1981},
  publisher={Taylor \& Francis}
}

@article{ghoshal2009random,
  title={Random hypergraphs and their applications},
  author={Ghoshal, Gourab and Zlati{\'c}, Vinko and Caldarelli, Guido and Newman, Mark EJ},
  journal={Physical Review E—Statistical, Nonlinear, and Soft Matter Physics},
  volume={79},
  number={6},
  pages={066118},
  year={2009},
  publisher={APS}
}

@article{10.1214/17-AOS1585,
author = {Rajarshi Mukherjee and Sumit Mukherjee and Subhabrata Sen},
title = {{Detection thresholds for the $\beta$-model on sparse graphs}},
volume = {46},
journal = {The Annals of Statistics},
number = {3},
publisher = {Institute of Mathematical Statistics},
pages = {1288 -- 1317},
year = {2018},
doi = {10.1214/17-AOS1585},
URL = {https://doi.org/10.1214/17-AOS1585}
}

@article{Benson-2018-simplicial,
 author = {Benson, Austin R. and Abebe, Rediet and Schaub, Michael T. and Jadbabaie, Ali and Kleinberg, Jon},
 title = {Simplicial closure and higher-order link prediction},
 year = {2018},
 doi = {10.1073/pnas.1800683115},
 publisher = {National Academy of Sciences},
 issn = {0027-8424},
 journal = {Proceedings of the National Academy of Sciences}
}

@inproceedings{10.5555/3305381.3305518,
author = {Guo, Chuan and Pleiss, Geoff and Sun, Yu and Weinberger, Kilian Q.},
title = {On calibration of modern neural networks},
year = {2017},
publisher = {JMLR.org},
booktitle = {Proceedings of the 34th International Conference on Machine Learning - Volume 70},
pages = {1321–1330},
numpages = {10},
location = {Sydney, NSW, Australia},
series = {ICML'17}
}

@article{hanley1982meaning,
  title={The meaning and use of the area under a receiver operating characteristic (ROC) curve},
  author={Hanley, James A. and McNeil, Barbara J.},
  journal={Radiology},
  volume={143},
  number={1},
  pages={29--36},
  year={1982}
}

@book{vanRijsbergen1979IR,
  title={Information Retrieval},
  author={van Rijsbergen, C. J.},
  year={1979},
  publisher={Butterworths},
  address={London}
}

@article{tony_cost_of_privacy,
author = {T. Tony Cai and Yichen Wang and Linjun Zhang},
title = {{The cost of privacy: Optimal rates of convergence for parameter estimation with differential privacy}},
volume = {49},
journal = {The Annals of Statistics},
number = {5},
publisher = {Institute of Mathematical Statistics},
pages = {2825 -- 2850},
year = {2021},
doi = {10.1214/21-AOS2058},
URL = {https://doi.org/10.1214/21-AOS2058}
}

@article{Stasi2014MF,
  title={$\beta$ models for random hypergraphs with a given degree sequence},
  author={Stasi, Despina and Sadeghi, Kayvan and Rinaldo, Alessandro and Petrovi{\'c}, Sonja and Fienberg, Stephen E},
  journal={arXiv preprint arXiv:1407.1004},
  year={2014}
}

@book{vershynin2018high,
  title={High-Dimensional Probability: An Introduction with Applications in Data Science},
  author={Vershynin, Roman},
  year={2018},
  publisher={Cambridge University Press},
  series={Cambridge Series in Statistical and Probabilistic Mathematics}
}

@article{FROSINI202197,
title = {New sufficient conditions on the degree sequences of uniform hypergraphs},
journal = {Theoretical Computer Science},
volume = {868},
pages = {97-111},
year = {2021},
issn = {0304-3975},
doi = {https://doi.org/10.1016/j.tcs.2021.04.006},
url = {https://www.sciencedirect.com/science/article/pii/S0304397521002103},
author = {Andrea Frosini and Christophe Picouleau and Simone Rinaldi}
}

@inproceedings{10.1007/978-3-642-37067-0_26,
author = {Frosini, Andrea and Picouleau, Christophe and Rinaldi, Simone},
title = {On the degree sequences of uniform hypergraphs},
year = {2013},
isbn = {9783642370663},
publisher = {Springer-Verlag},
address = {Berlin, Heidelberg},
url = {https://doi.org/10.1007/978-3-642-37067-0_26},
doi = {10.1007/978-3-642-37067-0_26},
booktitle = {Proceedings of the 17th IAPR International Conference on Discrete Geometry for Computer Imagery},
pages = {300–310},
numpages = {11},
location = {Seville, Spain},
series = {DGCI'13}
}

@article{rigollet2023highdimensionalstatistics,
      title={High-Dimensional Statistics}, 
      author={Philippe Rigollet and Jan-Christian Hütter},
      year={2023},
      journal={arxiv preprint: 2310.19244},
      archivePrefix={arXiv},
      primaryClass={math.ST},
      url={https://arxiv.org/abs/2310.19244}, 
}

@book{Tsybakov2009,
  author    = {Tsybakov, Alexandre B.},
  title     = {Introduction to Nonparametric Estimation},
  publisher = {Springer},
  address   = {New York, NY},
  year      = {2009},
  isbn      = {978-0-387-79052-7},
}

@inproceedings{acharya2021dpassouad,
  title     = {Differentially Private Assouad, Fano, and Le Cam},
  author    = {Acharya, Jayadev and Sun, Ziteng and Zhang, Huanyu},
  booktitle = {Proceedings of the 32nd International Conference on Algorithmic Learning Theory},
  series    = {Proceedings of Machine Learning Research},
  volume    = {132},
  pages     = {1--31},
  year      = {2021},
  editor    = {Feldman, Vitaly and Ligett, Katrina and Sabato, Sivan},
  publisher = {PMLR},
}

@article{social_groups_hypergraphs,
	author = {Palla, Gergely and Barab{\'a}si, Albert-L{\'a}szl{\'o} and Vicsek, Tam{\'a}s},
	date = {2007/04/01},
	doi = {10.1038/nature05670},
	id = {Palla2007},
	isbn = {1476-4687},
	journal = {Nature},
	number = {7136},
	pages = {664--667},
	title = {Quantifying social group evolution},
	url = {https://doi.org/10.1038/nature05670},
	volume = {446},
	year = {2007}}

@article{ji2016coauthorship,
  title={Coauthorship and citation networks for statisticians},
  author={Ji, Pengsheng and Jin, Jiashun},
  year={2016}
}

@article{karwa2016discussion,
  title={Discussion of ``Coauthorship and citation networks for statisticians"},
  author={Karwa, Vishesh and Petrovi{\'c}, Sonja},
  journal={The Annals of Applied Statistics},
  volume={10},
  number={4},
  pages={1827--1834},
  year={2016},
  publisher={JSTOR}
}

@article{karwa_et_al,
author = {Vishesh Karwa and Aleksandra Slavković},
title = {{Inference using noisy degrees: Differentially private $\beta$-model and synthetic graphs}},
volume = {44},
journal = {The Annals of Statistics},
number = {1},
publisher = {Institute of Mathematical Statistics},
pages = {87 -- 112},
year = {2016},
doi = {10.1214/15-AOS1358},
URL = {https://doi.org/10.1214/15-AOS1358}
}

@article{patania2017shape,
  title={The shape of collaborations},
  author={Patania, Alice and Petri, Giovanni and Vaccarino, Francesco},
  journal={EPJ Data Science},
  volume={6},
  number={1},
  pages={18},
  year={2017},
  publisher={Springer}
}

@article{grilli2017higher,
  title={Higher-order interactions stabilize dynamics in competitive network models},
  author={Grilli, Jacopo and Barab{\'a}s, Gy{\"o}rgy and Michalska-Smith, Matthew J and Allesina, Stefano},
  journal={Nature},
  volume={548},
  number={7666},
  pages={210--213},
  year={2017},
  publisher={Nature Publishing Group UK London}
}

@article{michoel2012alignment,
  title={Alignment and integration of complex networks by hypergraph-based spectral clustering},
  author={Michoel, Tom and Nachtergaele, Bruno},
  journal={Physical Review E—Statistical, Nonlinear, and Soft Matter Physics},
  volume={86},
  number={5},
  pages={056111},
  year={2012},
  publisher={APS}
}

@article{petri2014homological,
  title={Homological scaffolds of brain functional networks},
  author={Petri, Giovanni and Expert, Paul and Turkheimer, Federico and Carhart-Harris, Robin and Nutt, David and Hellyer, Peter J and Vaccarino, Francesco},
  journal={Journal of The Royal Society Interface},
  volume={11},
  number={101},
  pages={20140873},
  year={2014},
  publisher={The Royal Society}
}

@article{wang2013learning,
  title={On learning cluster coefficient of private networks},
  author={Wang, Yue and Wu, Xintao and Zhu, Jun and Xiang, Yang},
  journal={Social network analysis and mining},
  volume={3},
  number={4},
  pages={925--938},
  year={2013},
  publisher={Springer}
}

@article{chen2023private,
  title={Private estimation algorithms for stochastic block models and mixture models},
  author={Chen, Hongjie and Cohen-Addad, Vincent and d’Orsi, Tommaso and Epasto, Alessandro and Imola, Jacob and Steurer, David and Tiegel, Stefan},
  journal={Advances in Neural Information Processing Systems},
  volume={36},
  pages={68134--68183},
  year={2023}
}

@inproceedings{nguyen2016detecting,
  title={Detecting communities under differential privacy},
  author={Nguyen, Hiep H and Imine, Abdessamad and Rusinowitch, Micha{\"e}l},
  booktitle={Proceedings of the 2016 ACM on Workshop on Privacy in the Electronic Society},
  pages={83--93},
  year={2016}
}

@article{chakraborty2025primeprivacyawaremembershipprofile,
      title={PriME: Privacy-aware Membership profile Estimation in networks}, 
      author={Abhinav Chakraborty and Sayak Chatterjee and Sagnik Nandy},
      year={2025},
      journal={arxiv preprint: 2406.02794},
      archivePrefix={arXiv},
      primaryClass={stat.ME},
      url={https://arxiv.org/abs/2406.02794}, 
}

@article{pinot2018graph,
  title={Graph-based clustering under differential privacy},
  author={Pinot, Rafael and Morvan, Anne and Yger, Florian and Gouy-Pailler, Cedric and Atif, Jamal},
  journal={arXiv preprint arXiv:1803.03831},
  year={2018}
}

\appendix


\section{Proofs of results in Section \ref{sec:privacy_utility_tradeoff}}
\subsection{Proof of Theorem~\ref{thm:low_bound_local}}

To prove Theorem~\ref{thm:low_bound_local}, we invoke a locally private version of Fano’s inequality \cite[Corollary~3]{duchi2017minimaxoptimalprocedureslocally}. For convenience, we restate a simplified version below.

\begin{proposition}
\label{prop:modified_fano}
Suppose there exist $\gamma_1,\ldots,\gamma_J \in \mathbb R^{n}$ satisfying
$\|\gamma_j\|_\infty \le M$ for all $1 \le j \le J$. Consider the family of $r$-uniform $\beta$-models parameterized by $\gamma_1,\ldots,\gamma_J$, and consider the distributions of the privatized $r$-degree sequences of networks generated from these models obtained after applying a randomized mechanism $\mathcal M \in \mathfrak M^{\mathrm{loc}}_\varepsilon$. Let us denote such distributions by
$P^{\mathcal M}_{\gamma_1,\mathrm{deg}},\ldots,P^{\mathcal M}_{\gamma_J,\mathrm{deg}}$.
Assume that:
\begin{enumerate}
\item $\frac{1}{n}\|\gamma_i-\gamma_j\|_2^2 \ge 2s>0$ for all $i\neq j$,
\item For some $\alpha \in (0,1/8)$, we have
\[
\max_{\mathcal M \in \mathfrak M^{\mathrm{loc}}_{\varepsilon}}
\frac{1}{J^2}\sum_{i\neq j}
\mathrm{KL}\!\left(
P^{\mathcal M}_{\gamma_i,\mathrm{deg}},
P^{\mathcal M}_{\gamma_j,\mathrm{deg}}
\right)
\le \alpha \log J.
\]
\end{enumerate}
Then,
\[
\min_{\widehat\gamma\in \mathcal R^{\mathrm{loc}}_{\varepsilon,r}}
\max_{\|\gamma\|_\infty\le M}
\mathbb E\!\left[
\frac{1}{n}\|\widehat\gamma-\gamma\|_2^2\right]
\ge \frac{s}{2}\left\{1-\alpha-\frac{\log 2}{\log J}\right\}.
\]
\end{proposition}

\paragraph{Construction of the least favorable packing of the parameter set.}
To construct $\gamma_1,\ldots,\gamma_J$, we apply the Varshamov--Gilbert lemma
\cite[Lemma~2.9]{Tsybakov2009} to obtain vectors
$\omega_0,\ldots,\omega_J \in \{0,1\}^n$ such that
$J \ge 2^{n/8}$, $\omega_0=(0,\ldots,0)^\top$, and
\[
\|\omega_i-\omega_j\|_1 \ge \frac{n}{8}
\quad \text{for all } i\neq j.
\]
Define $\gamma_0=\omega_0$ and, for $1\le j\le J$,
\[
\gamma_j=\rho_n \omega_j,
\qquad
\rho_n := 4C\,(\varepsilon^2 n^{r-1})^{-1/2}.
\]
Then, for all $i\neq j$,
\[
\frac{1}{n}\|\gamma_i-\gamma_j\|_2^2
\ge \frac{2C}{\varepsilon^2 n^{r-1}},
\]
so condition (i) of Proposition~\ref{prop:modified_fano} holds.

\paragraph{Control of the Kullback--Leibler divergence.}
Fix an arbitrary mechanism $\mathcal M \in \mathfrak M^{\mathrm{loc}}_\varepsilon$.
By Theorem~1 of \cite{duchi2017minimaxoptimalprocedureslocally},
\[
\mathrm{KL}\!\left(P^{\mathcal M}_{\gamma_i,\mathrm{deg}},P^{\mathcal M}_{\gamma_j,\mathrm{deg}}\right)\le (e^\varepsilon-1)^2\mathrm{TV}^2\!\left(P_{\gamma_i,\mathrm{deg}},P_{\gamma_j,\mathrm{deg}}\right),
\]
where $P_{\gamma,\mathrm{deg}}$ denotes the distribution of the $r$-degree sequence generated under parameter $\gamma$.
Applying Pinsker’s inequality \cite[Lemma~4.9]{rigollet2023highdimensionalstatistics},
we obtain
\begin{align}
\label{eq:connect_kl_local}
\sum_{i\neq j}\mathrm{KL}\!\left(P^{\mathcal M}_{\gamma_i,\mathrm{deg}},P^{\mathcal M}_{\gamma_j,\mathrm{deg}}\right)
&\le (e^\varepsilon-1)^2\sum_{i\neq j}\mathrm{KL}\!\left(P_{\gamma_i,\mathrm{deg}},P_{\gamma_j,\mathrm{deg}}\right).
\end{align}
Since the $r$-degree sequence is a sufficient statistic for the $\beta$-model in \eqref{eq:r_uniform_beta_model},
\[
\mathrm{KL}\!\left(
P_{\gamma_i,\mathrm{deg}},
P_{\gamma_j,\mathrm{deg}}
\right)
=
\mathrm{KL}\!\left(
P_{\gamma_i},
P_{\gamma_j}
\right),
\]
where $P_\gamma$ denotes the distribution of the random $r$-uniform hypergraph under parameter $\gamma$.
Moreover, since $\varepsilon\in[0,\varepsilon_0]$, we have $(e^\varepsilon-1)^2\lesssim_{\varepsilon_0}\varepsilon^2$.
Retracing the proof of Theorem~2.2 of \cite{nandy_bhattacharya}, we obtain
\[
\mathrm{KL}\!\left(P_{\gamma_i},P_{\gamma_j}\right)\lesssim_{r}n^{r}\rho_n^2.
\]
Substituting into \eqref{eq:connect_kl_local} yields
\begin{align*}
\frac{1}{J^2}\sum_{i\neq j}\mathrm{KL}\!\left(P^{\mathcal M}_{\gamma_i,\mathrm{deg}},P^{\mathcal M}_{\gamma_j,\mathrm{deg}}\right)
&\lesssim \varepsilon^2 n^{r}\rho_n^2\;\lesssim\;C^2 n,
\end{align*}
uniformly over $\mathcal M \in \mathfrak M^{\mathrm{loc}}_\varepsilon$.

Choosing $C=C(\alpha,r,\varepsilon)$ sufficiently small ensures that condition (ii) of
Proposition~\ref{prop:modified_fano} holds with $\alpha\in(0,1/8)$. Since $n\ge 16$,
an application of Proposition~\ref{prop:modified_fano} completes the proof of
Theorem~\ref{thm:low_bound_local}.

\subsection{Proof of Theorem~\ref{thm:low_bound_central}}

Let $\mathcal V:=\{-1,+1\}^n$. For each $v\in\mathcal V$, define the parameter vector
\[
\beta_v:=\rho_n v,
\qquad 
\rho_n:=\frac{C_0}{\varepsilon n^{r-1}},
\]
where $C_0=C_0(r)>0$ is a constant to be chosen later. Under the assumption
$\varepsilon n^{r-1}\ge c$ (with $c>0$ absolute) and choosing $C_0\le cM$, we have
$\|\beta_v\|_\infty=\rho_n\le M$ for all $v\in\mathcal V$.
We work with the loss $\ell(\beta,\beta'):=\frac{1}{n}\|\beta-\beta'\|_2^2$.
For any $v,v'\in\mathcal V$,
\[
\ell(\beta_v,\beta_{v'})
=\frac{1}{n}\sum_{j=1}^n (\rho_n(v_j-v'_j))^2
=\frac{4\rho_n^2}{n}\, d_{\mathrm H}(v,v'),
\]
where $d_{\mathrm H}(\cdot,\cdot)$ denotes the Hamming distance on $\{-1,+1\}^n$.
Thus the Assouad separation condition holds with
\[
\ell(\beta_v,\beta_{v'}) \ge 2\tau\, d_{\mathrm H}(v,v'),
\qquad \text{where}\qquad \tau:=\frac{2\rho_n^2}{n}.
\]
For each $i\in[n]$, define the mixtures
\begin{align}
\label{eq:def_mixtures_fixed}
P_{+i}:=\frac{2}{|\mathcal V|}\sum_{v\in\mathcal V:\,v_i=+1} P_v,
\qquad
P_{-i}:=\frac{2}{|\mathcal V|}\sum_{v\in\mathcal V:\,v_i=-1} P_v,
\end{align}
where $P_v$ denotes the law of the $r$-uniform hypergraph $\beta$-model under parameter
$\beta_v$.

\paragraph{Coupling of $P_{+i}$ and $P_{-i}$.}
We construct a coupling $(G_X,G_Y)$ such that $G_X\sim P_{+i}$ and $G_Y\sim P_{-i}$, and
bound $\E[d_{\mathrm H}(G_X,G_Y)]$, where $d_{\mathrm H}$ is Hamming distance over all
$\binom{n}{r}$ hyperedge indicators.
Let $S_j \stackrel{\mathrm{i.i.d.}}{\sim}\mathrm{Unif}(\{-1,+1\})$ for $j\neq i$, and set
$S_{-i}:=\{S_j:\, j\neq i\}$. Define random parameter vectors $\beta_X,\beta_Y\in\R^n$ by
\[
(\beta_X)_j=
\begin{cases}
\rho_n S_j, & j\neq i,\\
\rho_n, & j=i,
\end{cases}
\qquad
(\beta_Y)_j=
\begin{cases}
\rho_n S_j, & j\neq i,\\
-\rho_n, & j=i.
\end{cases}
\]
By construction, $\beta_X$ has the law $P_v$ conditional on $v_i=+1$ (with the other
signs uniform), hence the induced hypergraph $G_X$ has marginal distribution $P_{+i}$; similarly
$G_Y$ has marginal distribution $P_{-i}$.

For each hyperedge $e\in\binom{[n]}{r}$, let $X_e:=\mathbf 1\{e\in G_X\}$ and $Y_e:=\mathbf 1\{e\in G_Y\}$.
If $i\notin e$, then $\sum_{j\in e}(\beta_X)_j=\sum_{j\in e}(\beta_Y)_j$, so the conditional distributions
of $X_e$ and $Y_e$ given $S_{-i}$ are identical; we couple them by setting $X_e=Y_e$ almost surely.

Now consider an edge $e$ with $i\in e$, and write $e=\{i\}\cup e'$ where $|e'|=r-1$.
Under the $r$-uniform $\beta$-model,
\[
\Pr(X_e=1\mid S_{-i})=\sigma\!\Big((\beta_X)_i+\sum_{j\in e'}(\beta_X)_j\Big)
=\sigma\!\Big(\rho_n + W_e\Big),
\]
\[
\Pr(Y_e=1\mid S_{-i})=\sigma\!\Big((\beta_Y)_i+\sum_{j\in e'}(\beta_Y)_j\Big)
=\sigma\!\Big(-\rho_n + W_e\Big),
\]
where
\[
W_e:=\sum_{j\in e'} \rho_n S_j
= \rho_n \sum_{j\in e\setminus\{i\}} S_j.
\]
We couple $(X_e,Y_e)$ by a maximal coupling of two Bernoulli random variables, which yields
\[
\Pr(X_e\neq Y_e\mid S_{-i}) = \big|\sigma(W_e+\rho_n)-\sigma(W_e-\rho_n)\big|.
\]
Since $\sigma'(t)=\frac{e^t}{(1+e^t)^2}\le \frac14$ for all $t\in\R$, the mean value theorem implies the
uniform bound
\[
\big|\sigma(w+\rho)-\sigma(w-\rho)\big|
\le 2\rho \cdot \sup_{t}\sigma'(t)
\le \frac{\rho}{2},
\qquad \forall w\in\R,\ \rho\ge 0.
\]
Applying this with $\rho=\rho_n$ gives
\[
\Pr(X_e\neq Y_e\mid S_{-i}) \le \frac{\rho_n}{2},
\qquad \text{hence}\qquad
\Pr(X_e\neq Y_e) \le \frac{\rho_n}{2}.
\]
Therefore, only hyperedges containing $i$ can disagree, and
\begin{align}
\label{eq:D_bound_fixed}
D
:=\E[d_{\mathrm H}(G_X,G_Y)]
&\le \sum_{e:\, i\in e} \Pr(X_e\neq Y_e)
\;\le\; \binom{n-1}{r-1}\cdot \frac{\rho_n}{2}
\;\le\; \frac{c_0'}{\varepsilon},
\end{align}
for an absolute constant $c_0'=c_0'(r,C_0)>0$ (using $\binom{n-1}{r-1}\asymp n^{r-1}$ and
$\rho_n=C_0/(\varepsilon n^{r-1})$). The same bound holds for every $i\in[n]$.

\paragraph{Apply DP--Assouad.}
By Theorem~3 of \cite{acharya2021dpassouad} (DP--Assouad with couplings), since $k=n$ and $\tau=2\rho_n^2/n$,
we obtain
\[
\min_{\widehat\beta \in \mathcal R^{\mathrm{cen}}_{\varepsilon,\delta,r}}
\max_{\|\beta\|_\infty\le M}
\E\!\left[\frac{1}{n}\|\widehat\beta-\beta\|_2^2\right]
\;\ge\;
\frac{n\tau}{2}\Big(0.9 e^{-10\varepsilon D}-10D\delta\Big)
\;=\;
\rho_n^2\Big(0.9 e^{-10\varepsilon D}-10D\delta\Big).
\]
Using $D\le c_0'/\varepsilon$ from \eqref{eq:D_bound_fixed}, we have $e^{-10\varepsilon D}\ge e^{-10c_0'}$.
Moreover, if $\delta=o(\varepsilon)$ (in particular, $\delta=o(1)$ suffices since $\varepsilon\le \varepsilon_0$ is fixed),
then $10D\delta \le 10(c_0'/\varepsilon)\delta = o(1)$. Choosing $C_0=C_0(r)$ sufficiently small (so that $c_0'$ is a fixed
constant) and taking $n$ large enough, the bracketed term is bounded below by an absolute constant.
Consequently, there exists $\mathfrak c_0=\mathfrak c_0(r,M,\varepsilon_0,\delta)>0$ such that
\[
\min_{\widehat\beta \in \mathcal R^{\mathrm{cen}}_{\varepsilon,\delta,r}}
\max_{\|\beta\|_\infty\le M}
\E\!\left[\frac{1}{n}\|\widehat\beta-\beta\|_2^2\right]
\;\ge\;
\mathfrak c_0\, \rho_n^2
\;=\;
\frac{\mathfrak c_0}{\varepsilon^2 n^{2(r-1)}}.
\]
Finally, by the (non-private) minimax lower bound (Theorem~2.2 of \cite{nandy_bhattacharya}),
\[
\min_{\widehat\beta}
\max_{\|\beta\|_\infty\le M}
\E\!\left[\frac{1}{n}\|\widehat\beta-\beta\|_2^2\right]
\;\ge\;
\frac{\mathfrak c_0}{n^{r-1}}.
\]
Combining the two lower bounds yields
\[
\min_{\widehat\beta \in \mathcal R^{\mathrm{cen}}_{\varepsilon,\delta,r}}
\max_{\|\beta\|_\infty\le M}
\E\!\left[\frac{1}{n}\|\widehat\beta-\beta\|_2^2\right]
\;\ge\;
\mathfrak c_0\cdot
\max\!\left\{\frac{1}{n^{r-1}},\,\frac{1}{\varepsilon^2 n^{2(r-1)}}\right\},
\]
which completes the proof.

\section{Proofs of results in Section \ref{sec:upper_bound_local}}
\subsection{Proof of Theorem~\ref{thm:local_diff_priv}}
The $\ell_1$-sensitivity of the $r$-degree sequence is defined as
\begin{align}
\Delta_1:=\max_{\substack{G,G'\\ |\mathcal E(G)\triangle\mathcal E(G')|=1}}\left\|d(G)-d(G')\right\|_1,
\end{align}
where $\mathcal E(G)$ and $\mathcal E(G')$ denote the hyperedge sets of $G$ and $G'$, respectively.

If $G$ and $G'$ differ by exactly one hyperedge, then the addition or removal of that hyperedge affects the degrees of exactly the $r$ vertices incident to it,
changing each of their $r$-degrees by one. Consequently,
\[
\left\|d(G)-d(G')\right\|_1 \le r,
\]
and this bound is attained. Therefore, the $\ell_1$-sensitivity of the $r$-degree sequence satisfies $\Delta_1=r$.

Given this sensitivity bound, adding independent discrete Laplace noise with parameter $e^{-\varepsilon/\Delta_1}$ to each coordinate of the degree sequence, as in \eqref{eq:local_priv_deg_seq}, yields an $\varepsilon$-locally differentially private release. The claim now follows directly from Lemma~1 of \cite{karwa_et_al}.

\subsection{Proof of Theorem \ref{thm:local_conv_rate}}
To prove Theorem~\ref{thm:local_conv_rate}, define the regularized negative log-likelihood function
\begin{align}
\label{eq:neg_loglik}
\ell^{\mathrm{loc}}_{\lambda,n}(\beta):=\sum_{(i_1,\ldots,i_r)\in{[n]\choose r}}
\log\left(1+e^{\beta_{i_1}+\cdots+\beta_{i_r}}\right)-\sum_{i=1}^n d^{\mathrm{loc}}_i\beta_i+\lambda\|\beta\|_2^2 .
\end{align}
The estimator in \eqref{eq:local_mle} can be equivalently written as
\[
\widehat\beta^{\mathrm{loc}}:=\argmin_{\beta\in\mathbb R^n}\ell^{\mathrm{loc}}_{\lambda,n}(\beta), \quad \text{such that $\|\wh \beta^{\mathrm{loc}}\|_\infty \le M$.}
\]
We shall use the following two auxiliary results.
\begin{lemma}
\label{lem:bound_gradient}
Let $\varepsilon\in(0,\varepsilon_0]$ and $\lambda\asymp n^{(r-1)/2}$. There exists a constant $C_1:=C_1(r,M)>0$ such that, for all
$\beta\in\mathcal B_{\infty,M}$,
\[
\|\nabla\ell^{\mathrm{loc}}_{\lambda,n}(\beta)\|_2^2
\;\le\;C_1\,\frac{n^{r}}{\varepsilon^2},
\]
with probability at least $1-e^{-c_1 n}$, for some constant $c_1>0$.
\end{lemma}

\begin{lemma}
\label{lem:bound_hessian}
Let $\varepsilon\in(0,\varepsilon_0]$ and suppose $\lambda\asymp n^{(r-1)/2}$. Then, for any $\gamma\in\mathcal B_{\infty,M}$, there exist constants
$0<C^{\mathrm{low}}_0\le C^{\mathrm{high}}_0<\infty$, depending only on $r$, and $M$, such that
\[
C^{\mathrm{low}}_0\,n^{r-1}
\;\le\;
\lambda_{\min}\!\left(\nabla^2\ell^{\mathrm{loc}}_{\lambda,n}(\gamma)\right)
\;\le\;
\lambda_{\max}\!\left(\nabla^2\ell^{\mathrm{loc}}_{\lambda,n}(\gamma)\right)
\;\le\;
C^{\mathrm{high}}_0\,n^{r-1}.
\]
In particular, the function $\ell^{\mathrm{loc}}_{\lambda,n}$ is strongly convex on $\mathcal B_{\infty,M}$ with modulus of order $n^{r-1}$.
\end{lemma}
Recall that $\widehat\beta^{\mathrm{loc}},\beta\in\mathcal B_{\infty,M}$. 
By first-order optimality,
\[
\nabla\ell^{\mathrm{loc}}_{\lambda,n}(\widehat\beta^{\mathrm{loc}})=0.
\]
For $t\in[0,1]$, define
\[
\bar\beta(t):=t\,\widehat\beta^{\mathrm{loc}}
+(1-t)\beta,\qquad \mbox{and} \qquad g(t):=(\widehat\beta^{\mathrm{loc}}-\beta)^\top
\nabla\ell^{\mathrm{loc}}_{\lambda,n}(\bar\beta(t)).
\]
Then $g(1)=0$ and
\(
g(0)=(\widehat\beta^{\mathrm{loc}}-\beta)^\top
\nabla\ell^{\mathrm{loc}}_{\lambda,n}(\beta),
\)
and $\wb \beta_t \in \mathcal B_{\infty,M}$ for all $t \in (0,1)$.
By the mean value theorem, there exists $t_0\in(0,1)$ such that $g(1)-g(0)=g'(t_0)$ so that $|g(0)|=|g'(t_0)|$.
A direct calculation gives
\[
g'(t_0)=(\widehat\beta^{\mathrm{loc}}-\beta)^\top
\nabla^2\ell^{\mathrm{loc}}_{\lambda,n}(\bar\beta(t_0))
(\widehat\beta^{\mathrm{loc}}-\beta).
\]
By the lower bound on the minimum eigenvalue of the Hessian in Lemma~\ref{lem:bound_hessian}, we have
\[
|g'(t_0)|\;\ge\;C^{\mathrm{low}}_0\,n^{r-1}
\|\widehat\beta^{\mathrm{loc}}-\beta\|_2^2.
\]
On the other hand, by Cauchy--Schwarz,
\[
|g(0)|
\le
\|\widehat\beta^{\mathrm{loc}}-\beta\|_2\,
\|\nabla\ell^{\mathrm{loc}}_{\lambda,n}(\beta)\|_2.
\]
Combining the two displays yields
\[
\|\widehat\beta^{\mathrm{loc}}-\beta\|_2
\;\le\;
\frac{1}{C^{\mathrm{low}}_0\,n^{r-1}}
\|\nabla\ell^{\mathrm{loc}}_{\lambda,n}(\beta)\|_2.
\]
Applying Lemma~\ref{lem:bound_gradient}, we obtain that with probability at least
$1-e^{-c_1 n}$,
\[
\|\widehat\beta^{\mathrm{loc}}-\beta\|_2
\;\lesssim_{r,M}\;
\sqrt{\frac{1}{\varepsilon^2 n^{r-2}}}.
\]
Define
\[
r^{\mathrm{loc}}_{n,\varepsilon}
:=
\widetilde C_0
\sqrt{\frac{1}{\varepsilon^2 n^{r-2}}},
\]
for a sufficiently large constant $\widetilde C_0=\widetilde C_0(r,M)>0$. Using $\|\widehat\beta^{\mathrm{loc}}-\beta\|_2^2\le 4M^2 n$ (by definition), we
decompose
\begin{align*}
\E\left[\|\widehat\beta^{\mathrm{loc}}-\beta\|_2^2\right]
&\le
\E\left[
\|\widehat\beta^{\mathrm{loc}}-\beta\|_2^2
\mathbbm 1\{\|\widehat\beta^{\mathrm{loc}}-\beta\|_2>r^{\mathrm{loc}}_{n,\varepsilon}\}
\right]\\
&\hskip 4em +
\E\left[
\|\widehat\beta^{\mathrm{loc}}-\beta\|_2^2
\mathbbm 1\{\|\widehat\beta^{\mathrm{loc}}-\beta\|_2\le r^{\mathrm{loc}}_{n,\varepsilon}\}
\right] \\
&\le
4M^2 n\,e^{-c_1 n}
+
\widetilde C_0^2\,\frac{1}{\varepsilon^2 n^{r-2}}.
\end{align*}
Dividing both sides by $n$ yields
\[
\frac{1}{n}
\E\left[\|\widehat\beta^{\mathrm{loc}}-\beta\|_2^2\right]
\;\le\;
C_0\,\frac{1}{\varepsilon^2 n^{r-1}},
\]
for a constant $C_0=C_0(r,M)>0$, which concludes the proof of
Theorem~\ref{thm:local_conv_rate}.

\subsection{Proof of Lemma \ref{lem:bound_gradient}}
Recall the regularized negative log-likelihood from \eqref{eq:neg_loglik},
where $d_i^{\mathrm{loc}}=d_i+Z_i$ and $Z_1,\ldots,Z_n$ are i.i.d.\ discrete Laplace
with parameter $\alpha=e^{-\varepsilon/r}$ (Lemma~1 of \cite{karwa_et_al}).
For $e\in {[n]\choose r}$ write $s_e(\beta):=\sum_{j\in e}\beta_j$ and
$p_e(\beta):=\exp(s_e(\beta))/(1+\exp(s_e(\beta)))$. For each $i\in[n]$,
\[
\bigl(\nabla \ell^{\mathrm{loc}}_{\lambda,n}(\beta)\bigr)_i
=
\sum_{e:i \in e} p_e(\beta) - d_i^{\mathrm{loc}} + 2\lambda \beta_i.
\]
Let $\mu_i(\beta):=\sum_{e: i \in e}p_e(\beta)$ and $\mu(\beta):=(\mu_1(\beta),\ldots,\mu_n(\beta))$.
Then
\begin{equation}
\label{eq:grad_decomp_final}
\nabla \ell^{\mathrm{loc}}_{\lambda,n}(\beta)
=
\bigl(\mu(\beta)-d\bigr) \;-\; Z \;+\; 2\lambda\beta,
\end{equation}
where $Z=(Z_1,\ldots,Z_n)$.
Hence, by the triangle inequality,
\begin{equation}
\label{eq:grad_triangle_final}
\|\nabla \ell^{\mathrm{loc}}_{\lambda,n}(\beta)\|_2
\le
\|\mu(\beta)-d\|_2 + \|Z\|_2 + 2\lambda\|\beta\|_2.
\end{equation}
Let $\mathcal V$ be a $(1/2)$-net of the Euclidean unit sphere $\mathbb S^{n-1}$.
By \cite[Lemma~5.2]{vershynin2018high}, $\log|\mathcal V|\le C n$ for an absolute constant $C>0$,
and for every $x\in\mathbb R^n$,
\begin{equation}
\label{eq:net_sphere_final}
\|x\|_2 \le 2\max_{v\in\mathcal V} v^\top x.
\end{equation}
Applying \eqref{eq:net_sphere_final} to $x=\mu(\beta)-d$ and to $x=Z$ yields
\begin{equation}
\label{eq:net_apply_final}
\|\mu(\beta)-d\|_2 \le 2\max_{v\in\mathcal V} v^\top(\mu(\beta)-d),
\qquad
\|Z\|_2 \le 2\max_{v\in\mathcal V} v^\top Z.
\end{equation}
Fix $\beta\in\mathcal B_{\infty,M}$.
Retracing the proof of Lemma~A.1 of \cite{nandy_bhattacharya}, there exist constants $c_{10},C_{10}>0$ (depending only on
$r$ and $M$ only), such that
\begin{equation}
\label{eq:model_fluct_final}
\max_{v\in\mathcal V} v^\top\bigl(d-\E_\beta[d]\bigr)
\le C_{10}\,n^{r/2},
\qquad
\text{with probability at least }1-e^{-c_{10}n}.
\end{equation}
Since $\E_\beta[d]=\mu(\beta)$, we have $\mu(\beta)-d=-(d-\E_\beta[d])$, and therefore
\begin{equation}
\label{eq:mu_minus_d_final}
\max_{v\in\mathcal V} v^\top\bigl(\mu(\beta)-d\bigr)
\le C_{10}\,n^{r/2},
\qquad
\text{with probability at least }1-e^{-c_{10}n}.
\end{equation}
For a discrete Laplace variable $Z$ with parameter $\alpha\in(0,1)$, the probability mass function in
Lemma~1 of \cite{karwa_et_al} implies the two-sided tail bound
\[
\Pr(|Z|\ge t)\le 2\alpha^{t}\qquad\text{for all }t\ge 0.
\]
Consequently $Z$ is sub-exponential and satisfies $\|Z\|_{\psi_1}\lesssim 1/\log(1/\alpha)$.
With $\alpha=e^{-\varepsilon/r}$ we have $\|Z\|_{\psi_1}\lesssim r/\varepsilon$.

Fix $v\in\mathbb S^{n-1}$. Since $Z_1,\ldots,Z_n$ are independent sub-exponential
with $\|Z_i\|_{\psi_1}\lesssim r/\varepsilon$ and $\|v\|_2=1$,
Corollary~2.9.2 of \cite{vershynin2018high} yields
\begin{equation}
\label{eq:bernstein_correct_final}
\Pr\!\left(\left|\sum_{i=1}^n v_i Z_i\right|>t\right)
\le
2\exp\!\left(
-c\,\min\left\{\frac{t^2\varepsilon^2}{r^2},\,\frac{t\varepsilon}{r}\right\}
\right)
\end{equation}
for an absolute constant $c>0$.
Taking $t=L(r/\varepsilon)n$ and choosing $L>0$ sufficiently large (depending only on $C$ in
$\log|\mathcal V|\le Cn$), we obtain from \eqref{eq:bernstein_correct_final} and a union bound over
$v\in\mathcal V$ that there exist constants $c_{20},C_{20}>0$ such that
\begin{equation}
\label{eq:noise_net_final}
\max_{v\in\mathcal V} v^\top Z \le C_{20}\,\frac{r}{\varepsilon}\,n,
\qquad
\text{with probability at least }1-e^{-c_{20}n}.
\end{equation}
Combining \eqref{eq:net_apply_final} and \eqref{eq:noise_net_final} yields
\begin{equation}
\label{eq:Z_norm_final}
\|Z\|_2 \le 2C_{20}\,\frac{r}{\varepsilon}\,n,
\qquad
\text{with probability at least }1-e^{-c_{20}n}.
\end{equation}
For $\beta\in\mathcal B_{\infty,M}$ we have $\|\beta\|_2\le M\sqrt n$, hence
\begin{equation}
\label{eq:ridge_term_final}
2\lambda\|\beta\|_2 \le 2\lambda M\sqrt n.
\end{equation}
Under the assumption $\lambda\asymp n^{(r-1)/2}$, the right-hand side of
\eqref{eq:ridge_term_final} is of order $n^{r/2}$.
On the event where \eqref{eq:mu_minus_d_final} and \eqref{eq:Z_norm_final} hold, combining
\eqref{eq:grad_triangle_final}, \eqref{eq:net_apply_final}, \eqref{eq:mu_minus_d_final},
\eqref{eq:Z_norm_final}, and \eqref{eq:ridge_term_final} gives
\[
\|\nabla \ell^{\mathrm{loc}}_{\lambda,n}(\beta)\|_2
\le
2C_{10}\,n^{r/2}
+2C_{20}\,\frac{r}{\varepsilon}\,n
+2\lambda M\sqrt n.
\]
Since $r\ge 2$ implies $n\le n^{r/2}$, and $\varepsilon\le \varepsilon_0$ implies $1\lesssim 1/\varepsilon$,
the above display yields
\[
\|\nabla \ell^{\mathrm{loc}}_{\lambda,n}(\beta)\|_2
\le
C\,\frac{n^{r/2}}{\varepsilon}
\]
for a constant $C=C(r,M,\varepsilon_0)>0$. Squaring both sides gives
\[
\|\nabla \ell^{\mathrm{loc}}_{\lambda,n}(\beta)\|_2^2
\le
C_1\,\frac{n^{r}}{\varepsilon^2}
\]
for a constant $C_1=C_1(r,M,\varepsilon_0)>0$.

Finally, taking a union bound over the two high-probability events in
\eqref{eq:mu_minus_d_final} and \eqref{eq:Z_norm_final} yields that the above inequality holds
with probability at least $1-e^{-c_1 n}$ for some constant $c_1>0$.
This completes the proof.

\subsection{Proof of Lemma \ref{lem:bound_hessian}}
By Lemma A.2 of \cite{nandy_bhattacharya},
\[
\frac{1}{4}\left(n^{r-1}e^{-r(M+\|\gamma-\beta\|_\infty)}+\lambda\right)
\;\le\;
\lambda_{\min}\!\left(\nabla^2\ell^{\mathrm{loc}}_{\lambda,n}(\gamma)\right)
\;\le\;
\lambda_{\max}\!\left(\nabla^2\ell^{\mathrm{loc}}_{\lambda,n}(\gamma)\right)
\;\le\;
\left(n^{r-1}+\lambda\right).
\]
The rest of the proof follows by observing that $\|\gamma\|_\infty, \|\beta\|_\infty \le M$ and $\lambda \asymp n^{(r-1)/2}$.

\section{Proofs of results in Section \ref{sec:central_privacy}}

\subsection{Proof of Theorem \ref{thm:central_privacy_valid}}

Choose the number of iterations $T \asymp \log n$ so that
\[
\varepsilon \le \varepsilon_0 \le \check{c}_1 T,
\]
for some absolute constant $\check{c}_1 > 0$. We first bound the $\ell_2$-sensitivity of the gradient $\nabla \ell_n(\beta)$.  
Define
\[
\Delta_2
:=
\max_{\substack{G,G'\\ |\mathcal E(G)\triangle\mathcal E(G')|=1}}
\left\|
\nabla \ell_n(\beta;G)-\nabla \ell_n(\beta;G')
\right\|_2,
\]
where $\nabla \ell_n(\beta;G)$ denotes the gradient computed using the degree
sequence induced by the hypergraph $G$. If $G$ and $G'$ differ by exactly one $r$-uniform hyperedge, then the addition or
removal of this hyperedge affects the degrees of exactly the $r$ incident
vertices, changing each of their degrees by one, while leaving all other
degrees unchanged. Consequently, the gradient differs in exactly $r$
coordinates, each by at most one in absolute value. Therefore,
\[
\left\|
\nabla \ell_n(\beta;G)-\nabla \ell_n(\beta;G')
\right\|_2
\le \sqrt{r},
\]
and this bound is tight. Hence the $\ell_2$-sensitivity satisfies
\[
\Delta_2 = \sqrt{r}.
\]
Each iteration of the algorithm releases a noisy gradient update obtained by
adding Gaussian noise proportional to $\Delta_2$, and thus implements a Gaussian
mechanism with $\ell_2$-sensitivity $\sqrt r$. By the design of the iterates
$\{\beta^t : t = 0,\ldots,T\}$ and the composition of $T$ such mechanisms, the
moment accountant method \cite[Theorem~1]{abadi_et_al} implies that the final
iterate $\beta^T$ is $(\varepsilon,\delta)$-edge differentially private. This completes the proof.

\subsection{Proof of Theorem \ref{thm:bound_central_privacy}}
By the inequality $\|a+b\|_2^2 \le 2\|a\|_2^2 + 2\|b\|_2^2$, we have
\begin{align}
\label{eq:master_equation}
\frac{1}{n}\mathbb E\!\left[\|\wh \beta^{\mathrm{cen}}-\beta\|^2_2\right]
\le
\frac{2}{n}\mathbb E\!\left[\|\wh \beta^{\mathrm{cen}}-\wh\beta\|^2_2\right]
+
\frac{2}{n}\mathbb E\!\left[\|\wh \beta-\beta\|^2_2\right].
\end{align}
Since $\wh\beta,\beta\in\mathcal B_{\infty,M}$, by Theorem~1 of \cite{nandy_bhattacharya}, we get
\[
\frac{1}{n}\mathbb E\!\left[\|\wh \beta-\beta\|^2_2\right]
\lesssim_M \frac{1}{n^{r-1}}.
\]
Thus it suffices to bound
\[
\frac{1}{n}\mathbb E\!\left[\|\wh \beta^{\mathrm{cen}}-\wh\beta\|^2_2\right]
=
\frac{1}{n}\mathbb E\!\left[\|\beta^T-\wh\beta\|^2_2\right].
\]
By \eqref{eq:get_strong_convexity_smooth}, the objective $\ell_n(\beta)$ is
$L$-smooth and $\mu$-strongly convex with
\[
L = \frac{1}{n}, \quad \mbox{and} \quad \mu = (4n)^{-1}e^{-2rM}.
\]
Define the unprojected iterate
\[
\wt \beta^{t}=\beta^{t}-\eta\bigl(\nabla\ell_n(\beta^{t}) + Z_t\bigr),
\]
where $Z_t$ is the injected Gaussian noise.
Since $\wh\beta\in\mathcal B_{\infty,M}$ and projection onto a convex set is non-expansive,
\begin{equation}
\label{eq:basic_update_step}
\|\beta^{t+1}-\wh\beta\|_2^2\le\|\wt\beta^{t}-\wh\beta\|_2^2 .
\end{equation}
Using Young’s inequality with parameter $\alpha = (1/16)e^{-4rM}$,
\begin{align}
\|\wt\beta^{t}-\wh\beta\|_2^2
&=
\bigl\|\beta^t-\wh\beta - \eta\nabla\ell_n(\beta^t) - \eta Z_t\bigr\|_2^2 \nonumber\\
&\le
\left(1+\frac{1}{32}e^{-4rM}\right)
\|\beta^t-\wh\beta - \eta\nabla\ell_n(\beta^t)\|_2^2
+
\bigl(1+32e^{4rM}\bigr)\eta^2\|Z_t\|_2^2 .
\end{align}
By strong convexity and smoothness,
\[
\nabla\ell_n(\beta^t)^\top(\wh\beta-\beta^t)
\ge
\mu\|\beta^t-\wh\beta\|_2^2,\qquad\|\nabla\ell_n(\beta^t)\|_2^2\le L^2\|\beta^t-\wh\beta\|_2^2.
\]
With the step size $\eta = 0.25\,ne^{-2rM}$, this yields the contraction
\begin{align}
\|\beta^{t+1}-\wh\beta\|_2^2
\le
\Bigl(1-\tfrac{1}{32}e^{-4rM}\Bigr)\|\beta^t-\wh\beta\|_2^2
+
(1+32e^{4rM})\|Z_t\|_2^2 .
\end{align}
Iterating the above inequality gives
\begin{align}
\|\beta^T-\wh\beta\|_2^2
&\le
\Bigl(1-\tfrac{1}{32}e^{-4rM}\Bigr)^T\|\wh\beta\|_2^2 \nonumber\\
&\quad
+
(1+32e^{4rM})
\sum_{k=0}^{T-1}
\Bigl(1-\tfrac{1}{32}e^{-4rM}\Bigr)^{T-k-1}
\|Z_k\|_2^2 .
\end{align}
Since $\|\wh\beta\|_2^2\le nM^2$ and $(1-x)^t\le e^{-xt}$,
choosing
\[
T= 32e^{4rM}\bigl\{(r-1)\log n + 2\log M\bigr\}
\]
ensures
\[
\Bigl(1-\tfrac{1}{32}e^{-4rM}\Bigr)^T \|\wh\beta\|_2^2
\le
n^{-(r-2)}.
\]
By Lemma~A.2 of \cite{tony_cost_of_privacy}, there exist constants
$\mathfrak C_1,\mathfrak c_1>0$ (depending only on $r$ and $M$) such that, with
probability at least $1-e^{-\mathfrak c_1 n}$,
\[
(1+32e^{4rM})
\sum_{k=0}^{T-1}
\Bigl(1-\tfrac{1}{32}e^{-4rM}\Bigr)^{T-k-1}
\|Z_k\|_2^2
\le
\mathfrak C_1
\frac{(\log n)\log(1/\delta)}{n^{2r-3}\varepsilon^2}.
\]
Combining the above,
\[
\|\beta^T-\wh\beta\|_2^2
\le
\mathfrak C_2
\max\!\left\{
\frac{1}{n^{r-2}},
\frac{(\log n)\log(1/\delta)}{n^{2r-3}\varepsilon^2}
\right\}
\]
with probability at least $1-e^{-\mathfrak c_1 n}$.
Since $\beta^T,\wh\beta\in\mathcal B_{\infty,M}$ almost surely,
$\|\beta^T-\wh\beta\|_2^2 \le 4nM^2$, and therefore
\[
\frac{1}{n}\mathbb E\!\left[\|\beta^T-\wh\beta\|_2^2\right]
\le
\mathfrak C_2\max\!\left\{\frac{1}{n^{r-1}},\frac{(\log n)\log(1/\delta)}{n^{2r-2}\varepsilon^2}
\right\}.
\]
Substituting this bound into \eqref{eq:master_equation} completes the proof.

\section{Description of the prediction metrics}
\label{sec:pred_mertics_def}

In Sections~\ref{sec:enron_email_dataset} and the subsequent section, we use the following metrics to evaluate the performance of hyperlink prediction.

\begin{enumerate}
    \item \textbf{ROC--AUC.} 
    Consider a binary prediction problem on a test set of candidate hyperlinks $\{y_e : e \in \mathcal E_{\mathrm{test}}\}$, where $y_e \in \{0,1\}$ indicates whether $e$ is a true hyperlink. Given an estimator $\widehat\beta$ (which may be any of $\widehat\beta_{\mathrm{MLE}}, \widehat\beta_{\mathrm{MLE},\lambda}, \widehat\beta^{\mathrm{loc}},$ or $\widehat\beta^{\mathrm{cen}}$), we assign to each $e \in \mathcal E_{\mathrm{test}}$ the score
    \[
    s_e(\widehat\beta) := \frac{\exp\!\left(\sum_{i \in e} \widehat\beta_i\right)}{1 + \exp\!\left(\sum_{i \in e} \widehat\beta_i\right)}.
    \]
    For a threshold $t \in [0,1]$, we classify $e$ as a predicted hyperlink if $s_e(\widehat\beta) \ge t$. The true positive rate (TPR) and false positive rate (FPR) are defined as
    \[
    \mathrm{TPR}(\widehat\beta,t) := \frac{\text{number of true positives}}{\text{number of true positives} + \text{number of false negatives}},
    \]
    and
    \[
    \mathrm{FPR}(\widehat\beta,t) := \frac{\text{number of false positives}}{\text{number of false positives} + \text{number of true negatives}}.
    \]
    The receiver operating characteristic (ROC) curve plots $\mathrm{TPR}(\widehat\beta,t)$ against $\mathrm{FPR}(\widehat\beta,t)$ as $t$ varies over $[0,1]$ \citep{hanley1982meaning}. The ROC--AUC ($\text{AUC}(\wh \beta)$) is defined as the area under this curve. Larger values of ROC--AUC indicate stronger separation between true hyperlinks and non-links.

    \item \textbf{Average precision (AP).} 
    The average precision score summarizes the precision--recall trade-off of a prediction rule and is particularly informative when evaluating performance on the positive class. For a sequence of thresholds $t_1,\ldots,t_K$, define the precision and recall as
    \[
    \mathrm{P}(\widehat\beta,t_k) := \frac{\text{number of true positives}}{\text{number of true positives} + \text{number of false positives}},
    \]
    \[
    \mathrm{R}(\widehat\beta,t_k) := \frac{\text{number of true positives}}{\text{number of true positives} + \text{number of false negatives}}.
    \]
    The average precision score (AP) is defined as \citep{vanRijsbergen1979IR}
    \[
    \mathrm{AP}(\widehat\beta) := \sum_{k=2}^K \bigl(\mathrm{R}(\widehat\beta,t_k) - \mathrm{R}(\widehat\beta,t_{k-1})\bigr)\,
    \mathrm{P}(\widehat\beta,t_k).
    \]
    Larger values of AP indicate better predictive performance.

    \item \textbf{Maximum F1-score.} 
    The F1-score \citep{vanRijsbergen1979IR} at threshold $t$ is defined as
    \[
    \mathrm{F1}(\widehat\beta;t) := 
    \frac{2\,\mathrm{P}(\widehat\beta,t)\,\mathrm{R}(\widehat\beta,t)}
    {\mathrm{P}(\widehat\beta,t) + \mathrm{R}(\widehat\beta,t)}.
    \]
    In sparse hyperlink prediction problems, the predicted probabilities $s_e(\widehat\beta)$ are often poorly calibrated, and using a fixed threshold may result in degenerate F1-scores (often equal to zero). To focus on ranking performance rather than calibration, we compute the F1-score over all distinct score values $\{s_e(\widehat\beta) : e \in \mathcal E_{\mathrm{test}}\}$ and report
    \[
    \mathrm{max\text{-}F1}(\widehat\beta) :=
    \max_{e \in \mathcal E_{\mathrm{test}}} \mathrm{F1}\bigl(\widehat\beta; s_e(\widehat\beta)\bigr).
    \]
    Larger values of the maximum F1-score indicate stronger ranking quality.

    \item \textbf{Expected calibration error (ECE).} 
    The expected calibration error (ECE) \citep{10.5555/3305381.3305518} measures the discrepancy between predicted probabilities and empirical outcome frequencies. Fix an integer $B \ge 1$ and partition the interval $[0,1]$ into $B$ disjoint bins $\{I_b\}_{b=1}^B$. For each bin $I_b$, define the index set $\mathcal I_b = \{ e \in \mathcal E_{\mathrm{test}} : s_e(\widehat\beta) \in I_b \}$, the empirical accuracy $\mathrm{acc}(b) = |\mathcal I_b|^{-1} \sum_{e \in \mathcal I_b} y_e$, and the average confidence $\mathrm{conf}(b) = |\mathcal I_b|^{-1} \sum_{e \in \mathcal I_b} s_e(\widehat\beta)$. The ECE is defined as
    \[
    \mathrm{ECE}(\wh \beta) := \sum_{b=1}^B \frac{|\mathcal I_b|}{n_{\mathrm{test}}}\,
    \bigl| \mathrm{acc}(b) - \mathrm{conf}(b) \bigr|.
    \]
    Smaller values of ECE indicate better calibration, while larger values reflect systematic over- or under-confidence in the predicted probabilities.
\end{enumerate}

\section{Numerical experiment: Cost of privacy in link prediction}

 \begin{table}[t]
\centering
\caption{Privacy cost in link prediction (differences vs non-private baselines)}
\label{tab:lp_privacy_cost}
\small
\setlength{\tabcolsep}{4pt}
\begin{tabular}{ccccc|ccc}
\toprule
$n$ & $\varepsilon$ & $\Delta$AUC$_\mathrm{loc}$ & $\Delta$F1$_\mathrm{loc}$ & $\Delta$ECE$_\mathrm{loc}$ & $\Delta$AUC$_\mathrm{cen}$ & $\Delta$F1$_\mathrm{cen}$ & $\Delta$ECE$_\mathrm{cen}$ \\
\midrule
50 & 0.01 & -0.209 & -0.047 & +0.001 & -0.153 & -0.043 & -0.000 \\
50 & 0.1 & -0.046 & -0.020 & +0.000 & -0.008 & -0.004 & -0.000 \\
50 & 1 & -0.001 & -0.000 & -0.000 & -0.000 & -0.000 & +0.000 \\
\midrule
100 & 0.01 & -0.117 & -0.044 & +0.000 & -0.038 & -0.019 & +0.000 \\
100 & 0.1 & -0.003 & -0.002 & -0.000 & -0.001 & -0.000 & +0.000 \\
100 & 1 & -0.000 & -0.000 & +0.000 & -0.000 & -0.000 & -0.000 \\
\midrule
150 & 0.01 & -0.045 & -0.022 & +0.000 & -0.010 & -0.005 & +0.000 \\
150 & 0.1 & -0.001 & -0.000 & +0.000 & -0.000 & -0.000 & +0.000 \\
150 & 1 & -0.000 & +0.000 & +0.000 & -0.000 & +0.000 & -0.000 \\
\midrule
200 & 0.01 & -0.018 & -0.009 & +0.000 & -0.004 & -0.002 & +0.000 \\
200 & 0.1 & -0.000 & -0.000 & -0.000 & -0.000 & -0.000 & +0.000 \\
200 & 1 & -0.000 & -0.000 & -0.000 & -0.000 & -0.000 & -0.000 \\
\bottomrule
\end{tabular}
\end{table}

We next investigate the cost of privacy for link prediction in $3$-uniform hypergraphs. We generate $\beta$ as in Section~\ref{sec:cost_param_est} and sample a hypergraph according to the $3$-uniform $\beta$-model in \eqref{eq:r_uniform_beta_model}. The observed hyperlinks are randomly split into a training set $\mathcal E_{\mathrm{obs}}$, containing $80\%$ of the edges, and a test set $\mathcal E_{\mathrm{test}}$. The test set consists of the remaining observed hyperlinks together with an equal number of randomly sampled non-edges. Using the same mechanisms and hyperparameters as in the previous section, we compute the estimators $\widehat\beta^{\mathrm{loc}}, \widehat\beta^{\mathrm{cen}}, \widehat\beta_{\mathrm{MLE},\lambda}$, and $\widehat\beta_{\mathrm{MLE}}$.  
For each candidate test edge $e=(i,j,k)\in\mathcal E_{\mathrm{test}}$, we form the corresponding link probability estimate $\widehat p_e$ by plugging the appropriate estimator into \eqref{eq:r_uniform_beta_model}.

We quantify the cost of privacy in link prediction by comparing the predictive performance of private estimators against their non-private counterparts. To this end, we consider the following performance gaps:
\begin{align}
    \Delta\text{AUC}_\mathrm{loc}&:= \text{AUC}(\widehat \beta^{\mathrm{loc}})-\text{AUC}(\widehat \beta_{\mathrm{MLE},\lambda}), \nonumber\\
    \Delta\text{AUC}_\mathrm{cen}&:= \text{AUC}(\widehat \beta^{\mathrm{cen}})-\text{AUC}(\widehat \beta_{\mathrm{MLE}}), \nonumber\\
    \Delta\text{F1}_\mathrm{loc}&:= \mathrm{max\text{-}F1}(\widehat \beta^{\mathrm{loc}})-\mathrm{max\text{-}F1}(\widehat \beta_{\mathrm{MLE},\lambda}), \nonumber\\
    \Delta\text{F1}_\mathrm{cen}&:= \mathrm{max\text{-}F1}(\widehat \beta^{\mathrm{cen}})-\mathrm{max\text{-}F1}(\widehat \beta_{\mathrm{MLE}}), \nonumber\\
    \Delta\text{ECE}_\mathrm{loc}&:= \mathrm{ECE}(\widehat \beta^{\mathrm{loc}})-\mathrm{ECE}(\widehat \beta_{\mathrm{MLE},\lambda}), \nonumber\\
    \Delta\text{ECE}_\mathrm{cen}&:= \mathrm{ECE}(\widehat \beta^{\mathrm{cen}})-\mathrm{ECE}(\widehat \beta_{\mathrm{MLE}}).
\end{align}

The results are summarized in Table~\ref{tab:lp_privacy_cost}. Overall, the cost of privacy is substantially more pronounced under local differential privacy than under central differential privacy. In both regimes, the magnitude of the privacy cost decreases as $n$ and $\varepsilon$ increase, consistent with improved parameter estimation in larger networks and under weaker privacy constraints. Finally, we observe that privacy has little effect on calibration: all methods—private and non-private—exhibit comparable levels of miscalibration, which is primarily driven by the sparsity of the generated hypergraphs rather than by the privacy mechanisms themselves.

\end{document}